\documentclass{article}

\usepackage[preprint]{neurips_data_2024}
\usepackage{amsmath, arydshln}
\usepackage{booktabs}
\usepackage{subcaption}
\usepackage{enumitem}
\usepackage{makecell}

\usepackage[toc,page]{appendix}
\usepackage[dvipsnames]{xcolor}
\usepackage{soul}
\usepackage{color, colortbl}
\usepackage[utf8]{inputenc} % allow utf-8 input
\usepackage[T1]{fontenc}    % use 8-bit T1 fonts
\usepackage{hyperref}       % hyperlinks
\usepackage{url}            % simple URL typesetting
\usepackage{booktabs}       % professional-quality tables
\usepackage{amsfonts}       % blackboard math symbols
\usepackage{nicefrac}       % compact symbols for 1/2, etc.
\usepackage{microtype}      % microtypography
\usepackage{xcolor}         % colors
\usepackage{titletoc}
\usepackage{xspace}
\usepackage{graphicx}
\usepackage{array} 
\usepackage{tcolorbox}

\newcommand{\name}{\textsc{\textbf{MUSE}}\xspace}
\usepackage{array, booktabs, xcolor, caption}
\usepackage{amsmath}
\usepackage{amssymb}
\usepackage{mathtools}
\usepackage{amsthm}
\usepackage{hhline}
\usepackage[capitalize,noabbrev]{cleveref}
\usepackage[affil-it]{authblk}

\usepackage{amssymb}
\usepackage{multirow}
\usepackage{pbox}
\usepackage{wrapfig}
\usepackage{subcaption}
\hypersetup{
    colorlinks=true,
	linkcolor=blue,
	filecolor=magenta,      
	urlcolor=blue,
	citecolor=blue,
}
\usepackage{cleveref}
\usepackage{bbding}
\usepackage{pifont}
\usepackage{wasysym}
\usepackage{amssymb}
\definecolor{amethyst}{rgb}{0.6, 0.4, 0.8}

\newcommand{\ensuretext}[1]{#1}
\newcommand{\marker}[2]{\ensuremath{^{\textsc{#1}}_{\textsc{#2}}}}
\newcommand{\arkcomment}[3]{\ensuretext{\textcolor{#3}{[#1 #2]}}}

\definecolor{lemon}{RGB}{255,247,0}
\definecolor{maize}{RGB}{250,237,94}
\definecolor{mustard}{RGB}{255,219,89}
\definecolor{ocre}{RGB}{241,103,35}
\definecolor{Tangerine}{RGB}{253,128,8}
\definecolor{framegreen}{RGB}{153, 188, 133}
\definecolor{bggreen}{RGB}{235, 250, 228}
\definecolor{c0}{cmyk}{1,0.3968,0,0.2588} 
\definecolor{c1}{cmyk}{0,0.6175,0.8848,0.1490} 
\definecolor{c2}{cmyk}{0.1127,0.6690,0,0.4431} 
\definecolor{c3}{cmyk}{0.3081,0,0.7209,0.3255} 
\definecolor{c4}{RGB}{164, 16, 52}
\definecolor{orange}{HTML}{E66100}
\definecolor{bluex}{HTML}{0C7BDC}
\definecolor{yellow}{HTML}{FFC20A}
\definecolor{lightpurple}{HTML}{E6E6FA}
\definecolor{lightbluee}{HTML}{e8f4f8}
\definecolor{blush}{rgb}{0.87, 0.36, 0.51}
\definecolor{c5}{HTML}{EE4E4E}
\definecolor{gggggg}{HTML}{EFEFEF}
\ifnum1=1
\newcommand{\chiyuan}[1]{\arkcomment{\marker{C}{Z}}{#1}{ForestGreen}}
\newcommand{\weijia}[1]{\arkcomment{\marker{W}{S}}{#1}{orange}}

\newcommand{\nascomment}[1]{\arkcomment{\marker{NA}{S}}{#1}{blue}}
\newcommand{\yang}[1]{\arkcomment{\marker{Y}{H}}{#1}{cyan}}
\newcommand{\jaechan}[1]{\arkcomment{\marker{J}{L}}{#1}{JungleGreen}}
\newcommand{\daogao}[1]{\arkcomment{\marker{D}{L}}{#1}{blue}}
\newcommand{\jz}[1]{\arkcomment{\marker{J}{Z}}{#1}{purple}}
\newcommand{\sadhika}[1]{\arkcomment{\marker{S}{M}}{#1}{blush}}

\newcommand{\task}[1]{\textcolor{red}{TODO: {#1}}}
\else
\newcommand{\chiyuan}[1]{}
\newcommand{\weijia}[1]{}
\newcommand{\nascomment}[1]{}
\newcommand{\yang}[1]{}
\newcommand{\jaechan}[1]{}
\newcommand{\daogao}[1]{}
\newcommand{\sadhika}[1]{}

\newcommand{\task}[1]{}
\fi

\usepackage{inconsolata}
\usepackage[T1]{fontenc}
\usepackage[scaled=0.95]{biolinum}

\usepackage{tcolorbox}
\definecolor{chart}{HTML}{1f77b4}

\newtcolorbox{example}[1][]{
  colback=chart!5!white,
  colframe=chart,
  floatplacement=floating,
  title=\centering \textsf{\small #1}
}
\newtcbox{\hlprimarytab}{on line, box align=base, colback=BlueGreen!20,colframe=blue,size=fbox,arc=3pt, before upper=\strut, top=-2.5pt, bottom=-4.5pt, left=-2pt, right=-2pt, boxrule=0pt}
\newtcbox{\hlsecondarytab}{on line, box align=base, colback=WildStrawberry!10,colframe=orange,size=fbox,arc=3pt, before upper=\strut, top=-2.5pt, bottom=-4.5pt, left=-2pt, right=-2pt, boxrule=0pt}
\newtcbox{\hlwhite}{on line, box align=base, colback=WildStrawberry!8,colframe=white,size=fbox,arc=2pt, before upper=\strut, top=-3pt, bottom=-4.5pt, left=-2pt, right=-2pt, boxrule=0pt}
\newtcbox{\hlyellow}{on line, box align=base, colback=BlueGreen!10,colframe=white,size=fbox,arc=2pt, before upper=\strut, top=-3pt, bottom=-4.5pt, left=-2pt, right=-2pt, boxrule=0pt}

\newcommand{\uashifted}{{\tiny$\uparrow$}}
\newcommand{\dashifted}{{\tiny$\downarrow$}}
\newcommand{\ua}[1]{{\scriptsize\hlsecondarytab{\uashifted{#1}}}}
\newcommand{\da}[1]{{\scriptsize\hlprimarytab{\dashifted{#1}}}}

\newcommand{\dar}[1]{{\scriptsize\hlsecondarytab{\dashifted{#1}}}}
\newcommand{\uar}[1]{{\scriptsize\hlprimarytab{\uashifted{#1}}}}

\newcommand{\orange}[1]{{\hlsecondarytab{#1}}}
\newcommand{\blue}[1]{{\hlprimarytab{#1}}}
\newcommand{\white}[1]{{\hlwhite{\textcolor{black}{\scriptsize\texttt{#1}}}}}
\newcommand{\yellow}[1]{{\hlyellow{\textcolor{black}{\scriptsize\textsf{#1}}}}}

\newcommand*{\affmark}[1][*]{\textsuperscript{\textnormal{#1}}}
\newcommand{\D}{\mathcal{D}}
\newcommand{\fu}{f_\textrm{unlearn}}

\newcommand{\fr}{f_\textrm{retrain}}
\newcommand{\ft}{f_\textrm{target}}
\newcommand{\fre}{f_\textrm{reinforce}}

\newcommand{\Dt}{\mathcal{D}_\textrm{train}}
\newcommand{\Du}{\mathcal{D}_\textrm{forget}}
\newcommand{\Dr}{\mathcal{D}_\textrm{retain}}
\newcommand{\Dh}{\mathcal{D}_\textrm{holdout}}
\newcommand{\GA}{\textsf{GA}\xspace}
\newcommand{\GDR}{\textsf{GDR}\xspace}
\newcommand{\KLR}{\textsf{KLR}\xspace}
\newcommand{\GD}{$\textsf{GA}_\GDR$\xspace}
\newcommand{\GKL}{$\textsf{GA}_\KLR$\xspace}
\newcommand{\NPO}{\textsf{NPO}\xspace}
\newcommand{\NPOD}{$\textsf{NPO}_\GDR$\xspace}
\newcommand{\NPOKL}{$\textsf{NPO}_\KLR$\xspace}
\newcommand{\TV}{\textsf{Task Vector}\xspace}
\newcommand{\WHP}{\textsf{WHP}\xspace}

\newcommand{\news}{\textsc{News}\xspace}
\newcommand{\books}{\textsc{Books}\xspace}

\usepackage{titlesec}
\titlespacing{\section}{0pt}{\parskip}{0pt}
\titlespacing{\subsection}{0.5\parskip}{5pt}{0pt}
\title{\raisebox{-6pt}{\includegraphics[scale=0.5]{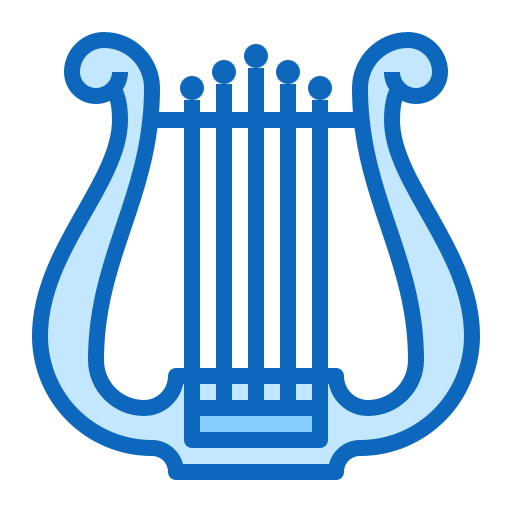}} \name:  Machine Unlearning Six-Way Evaluation for Language Models  }

\author{ 
\textbf{Weijia Shi$^*$}\affmark[1]\quad \textbf{Jaechan Lee$^*$}\affmark[1]\quad \textbf{Yangsibo Huang$^*$}\affmark[2]
\and\vspace{-6mm}
\textbf{Sadhika Malladi}\affmark[2]\quad\textbf{Jieyu Zhao}\affmark[3]\quad \textbf{Ari Holtzman}\affmark[4]\quad \textbf{Daogao Liu}\affmark[1]\quad 
\and\vspace{-6mm}
\textbf{Luke Zettlemoyer}\affmark[1]\quad
\textbf{Noah A. Smith}\affmark[1]\quad
\textbf{Chiyuan Zhang}\affmark[5]
\\
\vspace{-2mm}
\affmark[1]University of Washington\quad\affmark[2]Princeton University\\
\affmark[3]University of Southern California \quad
\affmark[4]University of Chicago
\quad\affmark[5]Google Research \\
\vspace{3mm}
\href{https://muse-bench.github.io}{\textcolor{magenta}{\texttt{https://muse-bench.github.io}}}
}

\begin{document}
\def\thefootnote{*}\footnotetext{Equal Contribution.}\def\thefootnote{\arabic{footnote}}
% \input{content/outline}
% \newpage

% \setcounter{page}{1}
\maketitle

    % \vspace{-15mm}
\begin{abstract}

Language models (LMs) are trained on vast amounts of text data, which may include private and copyrighted content, and data owners may request the removal of their data from a trained model due to privacy or copyright concerns. 
However, exactly unlearning only these datapoints (i.e., retraining with the data removed) is intractable in modern-day models, leading to the development of many approximate unlearning algorithms.
% As a result, machine unlearning has emerged as an active research area to address such requirements. 
% The perfect unlearning algorithm would be to retrain the model from scratch with the offending content  removed. 
% However, this is very costly, so various retraining-free algorithms have been proposed to approximate this ``perfect'' unlearning.
% Evaluation
%However, there are not systematic ways of comparing unlearning algorithms for various definitions of effectiveness prior to practical deployment. 
Evaluation of the efficacy of these algorithms has traditionally been narrow in scope, failing to precisely quantify the success and practicality of the algorithm from the perspectives of both the model deployers and the data owners.
We address this issue by proposing \name, a comprehensive machine unlearning evaluation benchmark that enumerates six diverse desirable properties for unlearned models: (1) no verbatim memorization, (2) no knowledge memorization, (3) no privacy leakage, (4) utility preservation on data not intended for removal, (5) scalability with respect to the size of removal requests, and (6) sustainability over sequential unlearning requests.
%both \textit{user} and \textit{deployer} expectations along six dimensions. Specifically, users typically have three main expectations regarding the unlearned models: (1) no verbatim memorization, (2) no knowledge memorization, (3) no privacy leakage. Deployers may consider three key metrics: (4) utility preservation on data not intended for removal, (5) scalability with respect to the size of removal requests, and (6) sustainability (no performance degradation under sequential unlearning requests). \name simulates real-world application scenario of unlearning copyright books and news articles.
Using these criteria, we benchmark how effectively eight popular unlearning algorithms on 7B-parameter LMs can unlearn Harry Potter books and news articles.
Our results demonstrate that most algorithms can prevent verbatim memorization and knowledge memorization to varying degrees, but only one algorithm does not lead to severe privacy leakage. 
Furthermore, existing algorithms fail to meet deployer's expectations, because they often degrade general model utility and also cannot sustainably accommodate successive unlearning requests or large-scale content removal.
Our findings identify key issues with the practicality of existing unlearning algorithms on language models, and we release our benchmark to facilitate further evaluations.\footnote{Our dataset and benchmark are available at \url{https://muse-bench.github.io}}

\end{abstract}

% unlearning is important and previous work on unlearning 
% but prev evaluation studies may not be systematic 
% start with an example (esp some settings that prev eval is not able to capture): author of harry potter would like to take down their work but harry pottern fans (e.g. who create wiki) may not want their contrinbution to be removed + other aspects (Sizes; may remove multiple books)
% standard: 1 verbmem, 2 knowmem, 3 privacy
% 4. remove books v.s. retain wiki (no side effect)
% 5. size varies (paragraph, charpter, book)
% 6. sequential (remove book1 ,book2, book3)

% let' say harry potter.. ..
% flow: p1 focsu on knownmem...in copryight, verbmem, even statistically privacy ... comprehensive mu unleanring should cover all the three 
% p2: there are some additional aspects ....
% Summary of our contributions

\section{Introduction}
\label{sec:intro}
% \weijia{add more citations}

\begin{figure}[t]
    % \vspace{-5mm}
    \centering
    \includegraphics[width=\linewidth]{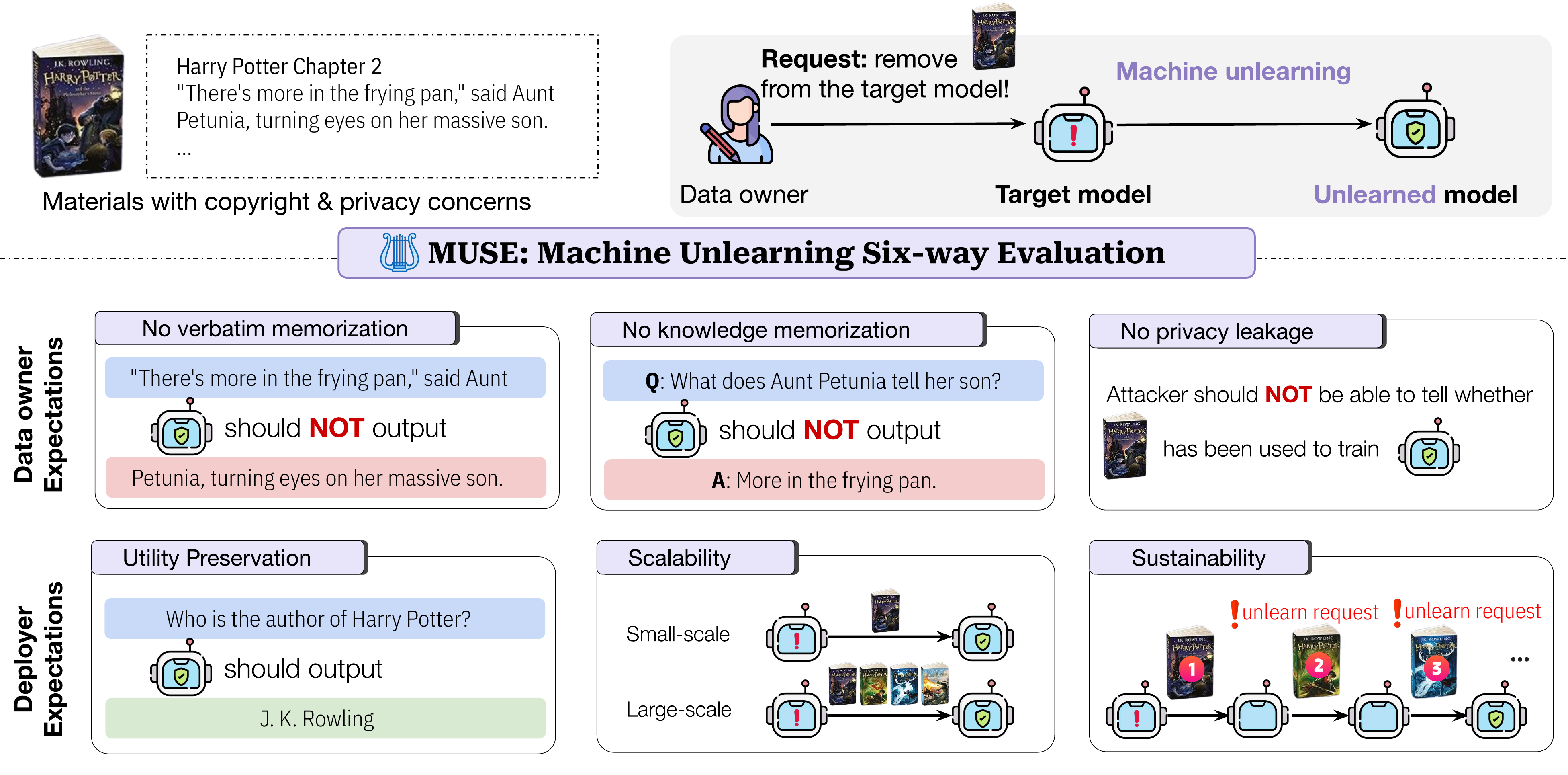}
    \caption{\textbf{\name evaluation focuses on six key dimensions of machine unlearning, addressing both \textit{data owner} and \textit{deployer} expectations.} For example, when an author (data owner) requests the unlearning of the Harry Potter books, they may expect the unlearned model to: (1) avoid generating verbatim copies of the text to protect copyright, (2) eliminate retention of factual knowledge from the books, and (3) not reveal whether the books were previously used in training to protect privacy.  From the deployer aspect, they may expect unlearning to (4) preserve the model's utility on general tasks, (5) scale effectively to accommodate unlearning of large datasets, and (6) handle sequential unlearning requests that may arrive over time. %\yang{change `user' to `data owner' in the figure}
    }
% \yang{Can be edited  \href{https://docs.google.com/drawings/d/1fZoWeNRXfBeKO1GMqPIt-U9jFCKFL1XUtyZg9hTMA_c/edit?usp=sharing}{here}.}
    \label{fig:main}
\end{figure}

 %\nascomment{suggest ``Language models (LMs)'' -- size of the models is not important to this paper's argument}  
Training language models (LMs) often involves using vast amounts of text data, which may inadvertently contain private and copyrighted content \citep{carlini2021extracting, henderson2023foundation,min2023silo, he2024fantastic}. 
In real-world applications, data owners may demand that their data be removed from a trained language model due to privacy or copyright concerns, as mandated for example by the %\jz{I may rewrite these 2 sentences to avoid being too similar to abstract.}
General Data Protection Regulation~\citep[GDPR,][]{GDPR2016a}.
%mandates that upon receiving a user’s request to opt out their data from training, model deployers must comply and remove the data from the model within 30 days (\textit{privacy compliance}). 
Moreover, recent copyright lawsuits \citep{githublitigation, openailawsuit2} 
%highlight the legal challenges faced when copyrighted content is used in model training and 
emphasize the need for removing copyrighted data from the model. %(\textit{copyright removal}).
%\weijia{change side effects to perserve utility}

These recent developments have intensified research interest in designing, evaluating, and improving \emph{machine unlearning} algorithms, which aim to transform an existing trained model into one that behaves as though it had never been trained on certain data~\citep{ginart2019making,liu2020federated, wu2020deltagrad, bourtoule2021machine, izzo2021approximate, gupta2021adaptive, sekhari2021remember, ye2022learning,ghazi2023ticketed}.
%As a result, machine unlearning, which 
%aims at the removal of specific training data from a trained model, has emerged as an important research problem~\citep{ginart2019making,liu2020federated, wu2020deltagrad, bourtoule2021machine, izzo2021approximate, gupta2021adaptive, sekhari2021remember, ye2022learning,ghazi2023ticketed}. 
%A simple algorithm that perfectly achieves the goal of unlearning is \emph{retraining}, which % on token level metrics~\citep{neurips-2023-machine-unlearning} 
Exact unlearning in LMs requires removing the undesired data (the \emph{forget set}) and retraining the model from scratch on the remaining data (the \emph{retain set}), which is too costly to be practical, especially for frequent unlearning operations.
As such, several efficient approximate unlearning algorithms have been proposed~\citep{eldan2023s,zhang2024negative}, but existing evaluations of LM unlearning 
on question answering~\citep{eldan2023s, Maini2024TOFUAT} cannot provide a holistic view of how practical and effective a particular unlearning algorithm is.
%have focused , but these benchmarks do not provide a holistic view of the efficacy of different unlearning algorithms.
% effective
In this work, we propose a systematic, multi-faceted framework called \name (\textbf{M}achine \textbf{U}nlearning \textbf{S}ix-Way \textbf{E}valuation; \S\ref{sec:benchmark}) to evaluate six desired properties for unlearning algorithms (\Cref{fig:main}).
Our criteria cover both the data owner's and the model deployer's desiderata for a practical unlearning algorithm.
Data owners require the LM to unlearn the precise tokens (\emph{verbatim memorization}), general knowledge encoded in the tokens (\emph{knowledge memorization}), and any indication that their data was included in the training set to begin with (\emph{privacy leakage}).
On the other hand, model deployers want to effectively accommodate many successive unlearning requests (\emph{sustainability}) on various sizes of forget sets (\emph{scalability}) without degrading the general model capabilities (\emph{utility preservation}).
We apply \name to evaluate \textbf{eight representative machine unlearning algorithms} (\S\ref{sec:method}) on \textbf{two datasets} (\S\ref{subsec:dataset}), focusing on the specific cases of unlearning Harry Potter books and news articles. 
Our findings indicate that most unlearning algorithms remove verbatim memorization and knowledge memorization with varying degrees of efficacy but operate at the cost of utility preservation and do not effectively prevent privacy leakage (\S\ref{subsec:results-users}).
In particular, negative preference optimization~\citep[\NPO;][]{zhang2024negative} and task vectors~\citep{ilharco2023editing} are especially effective in removing these types of memorization, but we find that \NPO often permits privacy leakage and both methods induce a sharp drop in the utility of the model.
Furthermore, testing their scalability and sustainability reveals that they both algorithms struggle with large forget sets and successive unlearning requests (\S\ref{subsec:results-deployer}). %\yang{Add pointers to result sections}

Our results highlight that unlearning algorithms generally fail to meet data owner expectations in preventing privacy leakage, which is one of the primary motivations for unlearning. 
Additionally, they struggle to meet all three of the aforementioned deployer expectations. 
Therefore, although it is increasingly desirable to find an efficient and effective unlearning algorithm amid rising concerns around privacy regulations and copyright litigations, our evaluation suggests that currently feasible unlearning methods are not yet ready for meaningful usage or deployment in real-world scenarios. These findings underscore the pressing need for further research in this area. We also release our benchmark to facilitate further evaluations and welcome extensions to other modalities.

\section{Machine Unlearning: Preliminaries and Notations}
Machine unlearning~\citep{ginart2019making, liu2020federated, izzo2021approximate, sekhari2021remember, gupta2021adaptive, ye2022learning, liu2024rethinking} 
% \yang{TODO: @weijia cite recent LLM applications} wu2020deltagrad, bourtoule2021machine,
has emerged as an 
important capability to accommodate 
data removal requirements that arise from scenarios with privacy or copyright concerns.
%We formulate machine unlearning as below. %and summarize the mathematical notation used throughout the paper in \Cref{tab:notation}.
%efficient alternative to full model retraining for addressing specific
%respecting privacy and removing copyrighted materials. 
%\chiyuan{Originally, this was written $\subset$ instead of $=$, we should choose one and agree upon what we call the ``retain set'' throughout the paper}\yang{Let's decide on this in our meeting.}. 
% \input{tables/notations}
% Consider a dataset $\Dt$ and a model $\ft$ trained on $\Dt$. 
% To simplify our notation, we consider the scenario of a single unlearning request, where a subset of the training data, i.e., the \emph{forget set} $\Du \subset \Dt$, is requested to be ``erased'' from $\ft$. We call the rest of the training set the \emph{retain set} ($\Dr = \Dt \setminus \Du$). 
% We also evaluate the model's behavior on a \emph{hold-out set} $\Dh$, drawn from the same distribution as $\Dt$ but disjoint from $\Dt$. The model $\ft$ has not been trained on $\Dh$.

We briefly describe the machine unlearning setting.
Consider a dataset $\Dt$ and a model $\ft$ trained on $\Dt$. 
Suppose we design an algorithm $\mathcal{U}$ to unlearn a specific subset (i.e., the \emph{forget set}) $\Du\subset\Dt$ from $\ft$.
We want to preserve performance on a \emph{retain set} $\Dr = \Dt \setminus \Du$, and we also  evaluate the model on an in-distribution but disjoint \emph{hold-out set} $\Dh$ which the model has never been trained on.
So, the unlearning algorithm $\mathcal{U}$ takes $\ft$, $\Du$, and, optionally, $\Dr$ and outputs an unlearned model $\fu$.
Exact unlearning ensures $\fu$ is behaviorally identical to the model resulting from retraining from scratch, denoted $\ft$, but such retraining is usually too costly in real world deployment, so we focus on evaluating approximate unlearning algorithms.

\section{The \name Evaluation Benchmark}
\label{sec:benchmark}

\begin{table}[t]
    \caption{\textbf{Comparison with a previous benchmark}: Unlike the previous benchmark TOFU~\citep{Maini2024TOFUAT}, which evaluates unlearning on synthetic Q\&A datasets, \name tackles real-world unlearning challenges: unlearning real-world large-scale corpus (22$\times$ larger) while taking into account six desiderata that are important to both data owners and deployers. More related works are discussed in Appendix \ref{app:related}.
    }
    \label{tab:comparison}
    \centering
    \setlength{\tabcolsep}{3pt}
    \resizebox{\linewidth}{!}{
    \begin{tabular}{clccc}
    \toprule
         & & \name (ours) & TOFU~\citep{Maini2024TOFUAT} \\
    \hline
     \multirow{6}{*}{\pbox{1.5cm}{\bf Evaluation criteria}}    &  C1. No verbatim memorization & $\checkmark$ \\
     &  C2. No knowledge memorization & $\checkmark$ &  $\checkmark$ \\
     &  C3. No privacy leakage  & $\checkmark$ \\
     &  C4. Utility preservation & $\checkmark$ &  $\checkmark$ \\
     &  C5. Scalability  & $\checkmark$ \\
     &  C6. Sustainability & $\checkmark$ \\
     \midrule
     \multirow{3}{*}{\pbox{1.5cm}{\bf Evaluation corpora}} & Domains &  \news and \books &  Synthetic autobiographies \\
     & Data Constitution & Verbatim text and knowledge set (Q \& A) &Q \& A \\ 
     & Scale ($\#$ tokens in forget set) & 0.8M for \news, 3.3M for \books& 0.15M\\
    \bottomrule
    \end{tabular}
    }
\end{table} \label{tbl:comparison}
% \yang{`Evaluation pipeline' or `evaluation benchmark' or `evaluation suite'? - benchmark}
\name evaluates a comprehensive 
% \nascomment{this word suggests that it's complete; no one could ever imagine adding another property to the list.  are we confident about that? (I think `comprehensive' also shows up elsewhere, like abstract/intro} 
set of desirable properties of machine unlearning across six facets. We detail the evaluation metrics in \S \ref{subsec:metric} and describe the evaluation corpus in \S \ref{subsec:dataset}.

\subsection{Evaluation Metrics}
\label{subsec:metric}
% Different from existing work that measure the effectiveness of unlearning methods through comparing certain easily computable statistics~\citep[e.g., loss, ][]{neurips-2023-machine-unlearning}, or performance~\citep[e.g., question answering,][]{eldan2023s, Maini2024TOFUAT} of the unlearned model on forget and retain sets,  we design our evalutions metrics to be very comprehensive  six dimensions and capture real-world  usablity in mind.
% \weijia{1: ideal behavior: cite the equation, 2: current methods, 3: user and deployer expectation}

Ideally, an unlearned model should behave as if it had never seen the forget set, exhibiting similar behavior to a retrained model on any corpus $\D$ such that $m(\fu, \D) \approx m(\fr, \D)$, where $m$ represents any evaluation metric. 
Prior 
%works focus on specific metrics like loss \citep[e.g.,][]{neurips-2023-machine-unlearning}, or 
evaluations on LM unlearning focus on
performance of specific tasks like question answering \citep[e.g.,][]{eldan2023s, Maini2024TOFUAT}.
% As shown in Table \ref{tbl:comparison}, prior benchmark focus on evaluating straightforward metrics such as question answering \citep[e.g.,][]{Maini2024TOFUAT}. 
%, which may not fully capture the data owner expectations and real-world deployment considerations of unlearning. 
However, these metrics do not faithfully reflect data owner expectations and real-world deployment considerations when performing unlearning.
To address this, we propose comprehensive evaluation metrics that consider both \textit{data owner} and \textit{deployer} expectations. A comparison between \name and the prior benchmark is shown in Table \ref{tbl:comparison}. 

\textbf{Data owner expectations.} 
When removing a forget set from a model, data owners typically have three main expectations regarding the unlearned model: 
(C1) \textbf{No verbatim memorization}: The model should not exactly replicate any details from the forget set. 
(C2) \textbf{No knowledge memorization}: The model should be incapable of responding to questions about the forget set. 
(C3) \textbf{No privacy leakage}: It should be impossible to detect that the model was ever trained on the forget set. 
For example, if a patient’s records are unlearned from a medical diagnosis model, in addition to verbatim and knowledge memorization checks, 
%the patient would expect the model not to reproduce the exact diagnosis or answer any questions related to it. 
it is also important that the patient's privacy is preserved -- we follow established practice in quantifying privacy using the membership inference test, which detects if a specific datapoint was used to train the model (\textit{member}), distinguishing it from non-training data (\textit{non-member})~\citep{shokri2017membership}.
In this case of unlearning a record from a diagnostic model, it is undesirable for the model to leak membership information, because it would be used to associate the patient with the disease. 
We quantify these data owner expectations with three evaluation metrics:

\begin{description}[leftmargin=16pt, itemsep=1pt]

%   The unlearned model should not reproduce the removed content verbatim. We define $\mathsf{VerbMem}(f, \mathcal{D})$ to quantify a model $f$'s verbatim memorization on dataset $\mathcal{D}$: for any given example $x \in \mathcal{D}$, we split it into a length-$l$ prompt $x_{[:l]}$ and its true continuation $x_{[l+1:]}$, %\nascomment{should be $x_{[l+1:]}$ I think (also below)}, 
%   and prompt the model $f$ to generate a continuation $f(x_{[:l]})$. The final verbatim memorization score is calculated by averaging the scores across all examples in $\mathcal{D}$:
% {\small\[ \mathsf{VerbMem}(f, \mathcal{D}) := \frac{1}{|\mathcal{D}|} \sum_{x \in \mathcal{D}} \mathsf{ROUGE}(f(x_{[:l]}), x_{[l+1:]})\]}
% where $\mathsf{ROUGE}(\cdot,\cdot)$ is the ROUGE-L F1 score~\citep{lin2004rouge}. An effectively unlearned model $\fu$ should satisfy $\mathsf{VerbMem}(\fu, \Du) \approx \mathsf{VerbMem}(\fr, \Du)$\yang{$\approx$ might be $\lessapprox$, see our discussion on slack.}, namely $\fu$ behaves similarly in terms of verbatim memorization to the retrained $\fr$ on the forget set $\Du$. 
\item[C1. No verbatim memorization] 
When a model has unlearned a medical record, it should not output its contents verbatim. 
We quantify the verbatim memorization $\mathsf{VerbMem}$ by prompting the model with the first $l$ tokens from a sequence $x_{[:l]} \in \Du$ and comparing the continuation outputted by the model $f$ to the true continuation $x_{[l+1:]} \in \Du$ using the ROUGE-L F1 score \citep{lin2004rouge}. 
%Specifically, we compare the model’s response $f(x_{[:l]})$ for each prompt $x_{[:l]}$ with its true continuation, using the ROUGE-L F1 score \citep{lin2004rouge}, and report the final verbatim memorization score as:
% The final verbatim memorization score is calculated by averaging the ROUGE-L F1 score \citep{lin2004rouge} across all examples in $\mathcal{D}$:
{\small
$$
\mathsf{VerbMem}(f, \mathcal{D}) := \frac{1}{|\mathcal{\Du}|} \sum_{x \in \mathcal{\Du}} \mathsf{ROUGE}(f(x_{[:l]}), x_{[l+1:]})
$$}
\item[C2. No knowledge memorization]
When a model has unlearned a medical record, it should no longer be able to answer questions about that record. 
We measure a model $f$’s memorization of knowledge from the forget set $\mathcal{\Du}$ as follows: for each example $x \in \mathcal{\Du}$ associated with a question-answer pair $(q, a)$,\footnote{Examples of question-answer pairs derived from the original corpus can be found in \Cref{tbl:dataset}.} we gather   the model's answer to the question $q$, denoted $f(q)$. 
We then average the ROUGE scores for all question-answer pairs in $\mathcal{\Du}$ to compute the knowledge memorization score $\mathsf{KnowMem}$:
 {\small\[\mathsf{KnowMem}(f, \mathcal{\Du}) := \frac{1}{|\mathcal{\Du}|} \sum_{(q, a) \in \mathcal{\Du}} \mathsf{ROUGE}(f(q), a)\]}

\begin{figure}[t]
    % \vspace{-14mm}
    \centering
    \includegraphics[width=\linewidth]{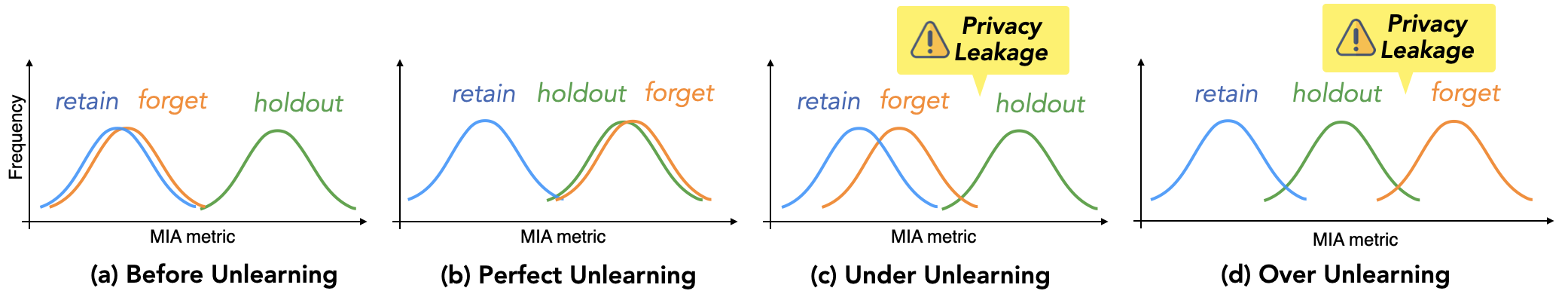}
    \vspace{-5mm}
    \caption{
    Distribution of the MIA metric (see C3) for $\Du$, $\Dh$, and $\Dr$. \textbf{Differences in the metric between forget and holdout sets indicate various unlearning outcomes of $\Du$, potentially leaking privacy.}
    A perfectly unlearned model (b) should show similar MIA metrics distribution for $\Du$ and $\Dh$. 
    % For a perfectly unlearned model (b), MIA metric distributions for $\Du$ and $\Dh$ should be similar. 
    Unlearning methods may fail by under-unlearning $\Du$, making it similar to $\Dr$ (c), or over-unlearning it, causing divergence from $\Dh$ (d).%\yang{@Weijia, let's update the colors to be consistent with Fig 3}
    }
    % Currnet unlearning may fail by insufficiently unlearning $\Du$, making it too close to the retain set $\Dr$ (c), or over-unlearning it, causing scores to diverge from $\Dh$ (d).}
    % The lower the MIA score For perfect unlearning, the MIA metric for $\Dh$ and $\Du$ should be similar and both  
    % Before unlearning, retain set and for the over unlearning setting, attacker infer that datapoints might have been used in training if they have very high MIA score
    
    \label{fig:mia}
\end{figure}
% \newpage
\item[C3. No privacy leakage] 
%The right to be forgotten allows the users to erase any influence of data on the model, preventing privacy leakage beyond the memorization tests. 
%One of the most established ways to quantify privacy risk is membership inference --- determining whether a specific data point was used to train the model~\citep{shokri2017membership}. 
%For example, a user might send unlearning requests for their clinical records to be removed from a model associated with a certain disease (e.g. for diagnosis), because the membership information could reveal that the patient carries the disease. 
%Therefore, it is desirable that the unlearned model no longer leaks membership information. 
As discussed previously, it is desirable that the unlearned model does not leak membership information indicating that $\Du$ was part of $\Dt$. To determine if a given example was used during training, \textit{membership inference attack} (MIA) exploits distributional differences in certain statistics (e.g., loss) between training (member) and non-training (non-member) data:
% A simple method for \emph{membership inference attack} (MIA) is loss thresholding: 
if the loss on the example is low, then it was likely used for training. 
As shown in \cref{fig:mia}, unlearning typically increases the loss on the example, but there are two possible ways that unlearning can fail to prevent privacy leakage: (1) \emph{under-unlearning}, when the loss is not made large enough; and (2) \emph{over-unlearning}, when the loss is made abnormally large.
%an unsuccessful algorithm could \emph{under-unlearn} if the loss is not raised high enough (compared to average unseen examples). 
%On the other hand, an overly aggressive algorithm could \emph{over-unlearn} and push the loss much higher above the normal range. 
%In both case, the membership privacy can be leaked. 
To accurately measure the privacy leakage, we employ Min-K\% Prob~\citep{shi2024detecting} 
% \footnote{Min-K\% Prob is used to distinguish between member (training) and non-member (non-training) examples by calculating the average log likelihood of minimum K\% tokens in a sentence}
, a state-of-the-art MIA method for LMs based on the loss, and compute the standard AUC-ROC score~\citep{murakonda2021quantifying,ye2022enhanced} of discriminating $\Du$ (members) and $\Dh$ (non-members).\footnote{An MIA algorithm compares its score to a given threshold to classify a given datapoint as a member or non-member. 
The AUC-ROC is a single value that summarizes the overall performance of the MIA algorithm by measuring its ability to discriminate between members and non-members across all possible thresholds.}
By comparing the AUC score with that of the retrained model, we define\footnote{Generally, $\textsf{AUC}(\fr; \Du, \Dh)\approx 0.5$, though sometimes there are intrinsic distribution shifts between $\Du$ and $\Dh$ that may bias the baseline away from 0.5.}
{\small $$\textsf{PrivLeak} := \frac{\textsf{AUC}(\fu; \Du, \Dh) - \textsf{AUC}(\fr; \Du, \Dh)}{\textsf{AUC}(\fr; \Du, \Dh)},$$}%
The \textsf{PrivLeak} metric for a good unlearning algorithm should be close to zero, whereas an over/under-unlearning algorithm will get a large positive/negative metric.
\end{description}

% \weijia{don't be too vague. we don't want the model to degrade over unelarnin grequests (sustantbailt)}
\textbf{Deployer expectations.} 
Model deployers have their own considerations for using unlearning algorithms in the real world.
Unlearning specific datapoints can unpredictably degrade model capabilities in ways that are difficult to recover. 
Moreover, deployers are expected to effectively accommodate somewhat large-scale forget sets and successive unlearning requests from data owners. As such, we
consider three key metrics: (C4) \textbf{utility preservation} on the retain set, (C5) \textbf{scalability} to handle large-scale content removal, and (C6) \textbf{sustainability} to maintain performance over sequential unlearning requests.
% We consider three other metrics crucial for deploying a machine unlearning. First, the unlearned model should \textbf{preserve utility} on the retain set. Second, deployers may receive requests to remove extensive content, such as the entire Harry Potter series, requiring \textbf{scalability}. Finally, if these requests come sequentially, 
% the method must not degrade over unlearning requests, requiring \textbf{sustainability}.
% Considering a request to unlearn the Harry Potter books in \Cref{fig:main}, model develoeprs may get unlearning request to unelarn entire harry potter erires (not just one book) therefore the method needs to be scalable, or the request comes sequentially, the model develeoprs have to apply the unelarning method seuqnetiallly.
% larger- \textit{scalable} to facilitate unlearning of large-scale datasets, and (6) be \textit{sustainable} in handling sequential unlearning requests.

% \ari{I don't think you make it clear enough what the difference between the user and the deployer is. It's a cute framing, so you could stick with it, but you'd have to reinforce it more, and you're introducing a lot of vocabulary. Another option would be to talk about ``unlearning metrics'' vs. ``usability metrics''.} \weijia{refer to the example...}

\begin{description}[leftmargin=16pt, itemsep=0.5pt]
    \item[C4. Utility preservation.] Model capabilities are often hard-won through expensive training procedures, so deployers would want an unlearning algorithm that preserves performance on the retain set. To quantify this, we evaluate the unlearned model's performance on the retain set using the knowledge memorization metric $\mathsf{KnowMem}(\fu, \Dr)$.

    \item[C5. Scalability.] We assess the scalability of unlearning methods by examining their performance on forget sets of varying sizes. Let $\mathcal{D}_u^c$ denote a forget set of size $c$, and $f_u^c$ be the corresponding unlearned model.
    For any data owner-valued metric such as utility preservation, we measure scalability by analyzing the trend of this metric as $c$ increases from small to large values.

    \item[C6. Sustainability.] Machine unlearning operations often need to be applied sequentially, as data removal requests may arrive at different times.\footnote{For example, under GDPR, if Alice requests the removal of her data and Bob submits another removal request 31 days later, both requests must be fulfilled within 30 days. This requires the model deployer to first unlearn Alice's data and then process Bob's request on the updated model.}
    We denote the unlearned model after processing the $k$-th request as $f_{u,k}$. To measure sustainability, we analyze the trend of any data owner-valued metric as the number of sequential unlearning requests $k$ increases.
    
    % For any user-valued metric $m$, such as $\mathsf{KnowMem}$ or $\mathsf{VerbMem}$ (computed using any dataset $\D \in \{\Dr, \Dh, \Du\}$), we measure sustainability by analyzing the trend of $m(\fuk, \D)$ as $k$ increases.
    %across various sequential unlearning stages $k$. \weijia{is this part clear?} \nascomment{it's ok but I'm not sure why we would average across $k$; wouldn't the first thing be to plot the metric on the $y$-axis, with $k$ on the $x$-axis?} \jaechan{I think we should just introduce the number of folds $K$ as a constant here (and perhaps mention that we set this to 4 during our experiments) and say e.g. across \textit{all} the unlearning requests $k \in [K]$.}\jaechan{Mention $\Du i$ and $\Du j$ are disjoint $\forall i \ne j$}
    % we evaluate the sustainability of an unlearning method by calculating the average over a collection of ${m(\fuk, \D)}$ across different $k$.
\end{description}

\subsection{Evaluation Corpus}
\label{subsec:dataset}

% \yang{WARNING: WIP!}

% \weijia{introduce how corpus looks like: original, qa (how to generate QA)}
% \weijia{talk about how to split the forget and retain, don't mention the purpose of the forget }

% \weijia{paragraph: news/HP: talk about corpus/forget and retain}

\name considers two representative types of textual data that may frequently involve unlearning requests: news articles~\citep{openailawsuit2} and books~\citep{eldan2023s}. 
These datasets are detailed as follows:
\begin{itemize}[leftmargin=16pt, nosep]
    \item \textbf{\news} consists of BBC news articles~\citep{li2023avoiding} collected after August 2023. All articles are randomly divided into (disjoint) forget, retain, and holdout sets.
    % The forget, retain and holdout set sets contain different dijoints sets of news articles.
    % \item \textbf{\books} consists of the Harry Potter book series. To simulate a real-world setting for testing side effects (\textbf{C4}), we intentionally include different types of materials in the forget and retain sets. The forget set contains the original books designated for unlearning, while the retain set contains related content from the Harry Potter FanWiki,\footnote{\url{harrypotter.fandom.com/wiki}} representing domain knowledge that the model should retain even after the original books have been unlearned.
    \item \textbf{\books} consists of the Harry Potter book series. To simulate a real-world setting for testing utility preservation (\textbf{C4}), we include different types of materials in the forget and retain sets. The forget set contains the original books, while the retain set contains related content from the Harry Potter FanWiki,\footnote{\url{harrypotter.fandom.com/wiki}} representing domain knowledge that should be retained after unlearning.

    % Notably, we deliberately keep \emph{different} types of materials in the forget and retain sets to simulate the real-world setting of testing for side effects (\textbf{C4}): The forget set includes the original books requested for unlearning, whereas the retain set contains related content from the Harry Potter FanWiki\footnote{\url{harrypotter.fandom.com/wiki}}, a simulation of domain knowledge which the model is expected to retain even after the original books are unlearned.
\end{itemize}

\newcolumntype{M}[1]{>{\centering\arraybackslash}m{#1}}
\begin{table}[t]
% \vspace{-14mm}
\centering
\small
\setlength{\tabcolsep}{7pt}
\renewcommand{\arraystretch}{1.15}
\caption{Examples of \name. Each corpus has \white{Verbatim} text and \yellow{Knowledge} sets (QA pairs derived from the original text) for evaluating verbatim and knowledge memorization. 
In \news, $\Du$ and $\Dr$ are two disjoint sets of news articles.
In \books, $\Du$ is the Harry Potter book series while $\Dr$ consists of wiki articles about the series. The sizes of the forget and retain sets are reported in tokens in \color{gray}{()}.
}
\resizebox{\linewidth}{!}{
\begin{tabular}{m{0.8cm}m{6.5cm}m{5.5cm}}
\toprule

\textbf{Corpus} & \textbf{Forget Set} & \textbf{Retain Set} \\
\hline
% \hline
\rowcolor{gggggg}
 & \textsc{\textbf{News Article}} \color{gray}{(0.8 M tokens)}  & {\textsc{\textbf{News Article}} {\color{gray}{(1.6 M tokens})}} \\

\multirow{4}{*}{\news} & \pbox{6.5cm}{\white{MP Stuart McDonald has been appointed as the SNP's } \\ \white{new treasurer} }  & \pbox{5.5cm}{\white{A father whose 12-year-old son was killed by} \white{an IRA bomb 30 years ago}} \\
\cmidrule{2-3}
 & \pbox{6.5cm}{\yellow{{\textbf{Q}: What position has Stuart McDonald MP been appointed to?}}\\\yellow{\textbf{A}: The SNP's new treasurer}} & \pbox{6.5cm}{\yellow{{\textbf{Q}: Who was affected by the IRA bomb 30 years ago?}}\\\yellow{\textbf{A}: A father whose 12-year-old son}}  \\
% \hline
\hline
\rowcolor{gggggg}
& {\textsc{\textbf{Harry Potter Books}} (\color{gray}{1.1 M tokens})} & {\textsc{\textbf{Harry Potter FanWiki}} (\color{gray}{{0.5 M tokens}})} \\
% \hline
\multirow{4}{*}{\books}  & \pbox{5.5cm}{\white{``There's more in the frying pan,'' said Aunt Petunia,}\\\white{turning eyes on her massive son.}} & \pbox{5.5cm}{\white{This page contains a list of spells:}\\\white{Portuguese for `open'.}} \\
\cmidrule{2-3}
 & \pbox{5.5cm}{\yellow{{\textbf{Q}: What does Aunt Petunia tell her son?}}\\ \yellow{\textbf{A}: There's more in the frying pan.}} & \pbox{5.5cm}{\yellow{{\textbf{Q}: What is the spell used to open things?}}\\\yellow{\textbf{A}: Portuguese}} \\
\bottomrule
\end{tabular}}
\label{tbl:dataset} 
\end{table}
For each corpus, we construct: 1) \white{\small Verbatim} text: the original text to assess the unlearning methods to remove verbatim memorization (\textbf{C1}), and 2) \yellow{\small Knowledge} set: a set of derived (question, answer) pairs
based on the original texts to evaluate the unlearning method's effectiveness in purging learned knowledge and preventing knowledge memorization (\textbf{C2}).
To create the Knowledge set, we partition the Verbatim text into excerpts and use GPT-4 \citep{openai2023gpt4} to generate (question, answer) pairs for each excerpt. 
% In particular, to construct the Knowledge set from the Verbatim text, we start by partition the Verbatim text into excerpts.
% Then, for each excerpt, we prompt GPT-4
% \citep{openai2023gpt4}
% to create a (question, answer) pair such that the answer is a verbatim extraction of the excerpt.
For more details about the dataset generation pipeline, see Appendix \ref{app:data_details}.
% \weijia{we use the orginal corpus to asses the verbatim meormozaion. further we generat knowledge sets ..}

% TODO: MOVE TO APPENDIX
% In particular, to construct the Knowledge set from the Verbatim text, we start by dividing the Verbatim text into 2048-token excerpts using LLaMA-2's tokenizer.
% For each (question, answer) pair we need to generate, we randomly select an excerpt from this set and prompt GPT-4 to create a JSON object with two fields: ``question'' (a question that can only be answered using specific information from the excerpt) and ``answer'' (an answer to the ``question'' extracted verbatim from the excerpt).
% We exclude any pairs whose answers cannot be found verbatim in their corresponding excerpts.
% This verbatim requirement ensures that the "knowledge" being tested is precisely the model's ability to correctly associate each question with the relevant part of the training data and nothing else.
% This approach is crucial for evaluating the model's ability to correctly interpret questions while also being applicable to our copyright-related tasks, where generating a verbatim extraction is more concerning than producing its paraphrase.
\Cref{tbl:dataset} provides  examples from the news and books corpora.
The details of the dataset splits and dataset sizes are provided in \Cref{app:data_details}. 

\section{Unlearning Methods}
\label{sec:method}
% \weijia{can someone check whether the current method section is clear? TAT}\yang{I can make a pass later tonight (6/3)!!}
% We evaluate 8 representative unlearning methods covering 4 families methods:
% \textbf{Four families of methods}
% \begin{itemize}
%     \item GA
%     \item NPO
%     \item TV
%     \item WHP
% \end{itemize}

% \textbf{Two regualrizers based on retain set. (optional)} 
% \begin{itemize}
%     \item GDR
%     \item KLR
% \end{itemize}

% We evaluate 8 representative unlearning methods covering 4 families methods:
We evaluate eight efficient approximate unlearning methods belonging to four families of algorithms. 
%These methods aim to efficiently
%adjust the model so that it behaves as if it has never seen the forget set, 
%unlearn
%without the need for full retraining.

% The objective of these methods is to effectively achieve a scenario in which the model has never been exposed to the forget set, without the need for complete retraining.

% We experiment with eight unlearning methods. The goal of these methods is to make the model forget the forget set while keeping good performance on the retain set. (achieve the counterfactual of “not having seen
% the forget set” without retraining.)

\textbf{Four families of unlearning methods.}
We first introduce four families of unlearning methods, which serve as the basis for the eight methods we evaluate.

\begin{itemize}[leftmargin=16pt, itemsep=1pt]
\item \textbf{Gradient Ascent} (\GA) minimizes the likelihood of correct predictions on $\Du$ by performing gradient ascent on the cross-entropy loss (the opposite of conventional learning with gradient descent).
\GA has achieved mixed results: while \cite{jang-etal-2023-knowledge} found it effective for unlearning examples from the Enron email dataset~\citep{klimt2004enron} with minimal performance degradation, \cite{ilharco2023editing} reported that \GA significantly harms general model utility when unlearning a high-toxicity subset of the Civil Comments dataset~\citep{borkan2019nuanced}.
%For example, \cite{jang-etal-2023-knowledge} applied \GA to the Enron email dataset~\citep{klimt2004enron} and found it effective for unlearning with minimal performance degradation.
%Conversely, \cite{ilharco2023editing} used \GA on a high-toxicity subset of the Civil Comments \citep{borkan2019nuanced} dataset and reported that \GA degrades the general utility of the model to an ``unacceptable level.''
% The cost of \GA as an unlearning method is the time it takes for the model to achieve a sufficiently high loss on $\Du$.

\item \textbf{Negative Preference Optimization} (\NPO; \citealp{zhang2024negative}) treats the forget set as negative preference data and adapts the offline DPO objective~\citep{rafailov2023direct} to tune the model to assign low likelihood to the forget set without straying too far from the original model $\ft$.
%makes the prediction probability on $\Du$ as small as possible by minimizing the following loss on $\Du$:
% so that the prediction probability on $\Du$ is as small as possible:
{\small
\begin{align*}
\mathcal{L}_{\mathrm{NPO}} (\theta)=-\frac{2}{\beta} \mathbb{E}_{x \sim \Du} \left[\log \sigma\left(-\beta \log \frac{f_\theta(x)}{\ft (x)}\right)\right],
\end{align*}
}where $f_\theta$ refers to the model that undergoes unlearning, $\sigma$ is the sigmoid function, and $\beta$ is a hyperparameter that controls the allowed divergence of $f_\theta$ from its initialization $\ft$.
Following \cite{rafailov2023direct,zhang2024negative}, we fix $\beta = 0.1$ in our experiments.
% \chiyuan{I changed $\sigma$ to $\mathsf{Sigmoid}$ so that we do not need to separately explain it, but please double check it is correct}
% \weijia{define what is x, y. they are sampled from $\Du$?}
% \weijia{i don't think we need to explain how it is related to DPO... just explain more clearly what this loss function does}
% This approach modifies the loss function of direct preference optimization \citep{rafailov2023direct} by omitting a term involving the winning sample, based on the intuition that samples from $\Du$ are always undesirable.
% Like \GA, the cost of \NPO is the time it takes for the model to achieve a high loss on $\Du$.

\item \textbf{Task Vectors} (\citealp{ilharco2023editing}) derived from straightforward arithmetic on the model weights can effectively steer neural network behavior.
We adapt task vectors to perform unlearning in two stages.
First, we train
% \nascomment{be more specific:  you continue training it on the data?  until when?  or do you train from scratch on $\Du$?}
$\ft$ on $\Du$ until the model overfits, yielding a reinforced model $\fre$.
We then obtain a task vector related to $\Du$ by calculating the weight difference between $\ft$ and $\fre$. 
% thus specifying a direction of overfitting in the weight space \weijia{change a word, don't call overfitting}.
To achieve unlearning, we subtract this task vector from $\ft$'s weights, intuitively moving the model away from the direction it used to adapt to $\Du$ -- i.e., $\fu = \ft - (\fre - \ft)$.
%, i.e.
%$\fu = \ft - (\fre - \ft)$.
% The cost of \TV is equivalent to creating the reinforced model by overfitting $\ft$ to $\Du$.

% \nascomment{confused why we're referencing a subsection here}
% \footnote{
% Relabeling the named entities is a crucial part of their main algorithm.% We skip its implementation because not every dataset (e.g., BBC) has a straightforward set of named entities like in Harry Potter books. \yang{This footnote seems to be a bit out of scope? It seems that we don't have the `relabeling' sentence in the main text.} } 
% {\small $$v_{\text {target}}-\alpha \operatorname{ReLU}\left(v_{\text {reinforced}} - v_{\text {target}}\right),$$}
% achieves unlearning by directly sampling from an interpolated output distribution between $\ft$ and $\fre$ on-the-fly. Let's $p(f)$ denotes the output distribution from the model $f$. Specifically, the methods sample from
% $\sim p(\ft)- \alpha(p(\fre) - p(\ft))$
% where $v_{\text{target}}$ is the output logit of $\ft$, $v_{\text{reinforced}}$ that of the reinforced model, and $\alpha$ is a hyperparameter that we fix as 5 during our evaluation. When generating each token, the model samples from this interpolated distribution instead of the original output distributions of $\ft$ or $\fre$. This approach allows the model to generate text that partially retains the knowledge of $\ft$ while also being influenced by the unlearning achieved through $\fre$, without the need for explicitly interpolating the model weights.
\item \textbf{Who's Harry Potter} (\WHP; \citealp{eldan2023s}) defines the unlearned model $\fu$ as the interpolation between the target model $\ft$ and the reinforced model $\fre$. 
Let $p_f(\cdot|x)$ denote the token distribution parametrized by the model $f$ when given a prompt $x$ as input.
Then, concretely, for any input $x$, \WHP samples the next token from 
$$p_{\fu}(\cdot| x) = p_{\ft}(\cdot | x) - \alpha(p_{\fre}(\cdot | x) - p_{\ft}(\cdot | x))$$ 
%$$y \sim p(\ft) - \alpha(p(\fre) - p(\ft))$$
where $\alpha$ is a hyperparameter that controls the interpolation between the two models. 
% where $\alpha$ is weight adjustment. When generating each token, the model samples from this interpolated distribution ($p_{\text{interpolated}}$) instead of the original output distributions of either $\ft$ or $\fre$

% consider the same reinforced model $\fre$ as above that is overfit to $\Du$.
% Rather than manipulating model parameters, they interpolate the output distribution between $\ft$ and $\fre$ on-the-fly as follows:
% {\small $$v_{\text {target}}-\alpha \operatorname{ReLU}\left(v_{\text {reinforced}} - v_{\text {target}}\right),$$}

% where $v_{\text{target}}$ is the output logit of $\ft$, $v_{\text{reinforced}}$ that of the reinforced model, and $\alpha$ is a hyperparameter that we fix as 5 during our evaluation.
% Unlearning is achieved by sampling from this interpolation.
% The cost of \WHP is that of creating the reinforced model by overfitting $\ft$ to $\Du$ in addition to the space and time overheads caused by storing and running both the target and reinforced models.

\end{itemize}
\textbf{Two regularizers for utility preservation.}
GA and NPO are not explicitly designed for utility preservation, so we discuss several regularization strategies that either improve the performance on the retain set or ensure the unlearned model remains close to the target model during unlearning.

\begin{itemize}[nosep, leftmargin=16pt]

% This ensures that the unlearned model still achieves descent perofrmance on retian set 
% the target model only unlearns the necessary components of $\Du$ without sacrificing its overall utility as a language model. 
% This approach can be applied to both \GA and \NPO.
% \nascomment{I think you mean and the unlearned model here}

% \item \textbf{KL Divergence Minimization on the Retain Set} (\KLR;
% \citealp{Maini2024TOFUAT, zhang2024negative})   regularizes the unlearning objective with the Kullback-Leibler divergence of the token logits on $\Dr$ under target model $\ft$ and
% the model undergoing unlearning $\fu$. By minimizing the KL divergence, \KLR aims to ensure that the output distribution of the unlearned model closely approximates the output distribution of the target model on the retain set. 
\item \textbf{Gradient Descent on the Retain Set} (\GDR;
\citealp{liu2022continual, Maini2024TOFUAT, zhang2024negative}) augments the unlearning objective with a standard gradient descent learning objective on the cross-entropy of the retain set $\Dr$ to more directly train the model to maintain its performance on $\Dr$.
%incorporates a gradient descent term with respect to the retain set $\Dr$ into the unlearning objective function. 
% $\fu$ on $\Dr$, this approach ensures that the unlearned model preserves its performance on the retained data while undergoing the unlearning process.  

\item \textbf{KL Divergence Minimization on the Retain Set} (\KLR;
\citealp{Maini2024TOFUAT, zhang2024negative}) encourages the unlearned model's probability distribution $p_{\fu}(\cdot | x)$ to be close to the target model's distribution $p_{\ft}(\cdot | x)$ on inputs from the retain set $x\in\Dr$. 
\end{itemize}
% Both of these regularizers increase the cost of unlearning by requiring an additional forward pass of the model on a sample from $\Dr$ for each step.

\textbf{List of methods.}
We combine \GA and \NPO with the two regularizers $\GDR$ and $\KLR$,\footnote{
These regularizers are not compatible with \TV and \WHP, because \TV involves purposefully overfitting a model to $\Du$ when deriving the task vector, and \WHP is a test-time technique where the unlearning operation involves no optimization by itself.
} which yields four new combinations.
Hence, we end up with a total of 8 candidate unlearning methods:
\GA, \GD, \GKL, \NPO, \NPOD, \NPOKL, \TV, and \WHP. 
In general, the cost of the approximate unlearning method is negligible compared to retraining. Details about the efficiency of these methods are reported in Appendix \ref{app:efficiency}.

% \nascomment{missing:  this section should have some discussion about the relative costs of these methods in terms of training and runtime/inference.  you made a big thing about how retraining is costly; ideally we would quantify all of these against retraining, in terms of computational cost.  oh wait, I guess that's part of scalability?  still some discussion of the base cost of each of these methods seems important here}
\section{Experiments}
\label{sec:exp}

We evaluate the eight representative unlearning methods using the experimental setup described in \S \ref{subsec:exp_setup}. We present the results for data owner expectations in \S \ref{subsec:results-users} and for deployer expectations in \S \ref{subsec:results-deployer}.

\subsection{Experimental Setup}
\label{subsec:exp_setup}
\begin{table}[t]
\vspace{-8mm}
\caption{
\textbf{Most unlearning methods effectively remove verbatim and knowledge memorization but significantly impact utility and privacy.}
We evaluate the 8 algorithms described in \S\ref{sec:method} on 4 of the criteria in \name. 
We include the results of $\fr$ for reference.
%$\fr$ is the model retrained without the forget set, so it benchmarks the performance after exact unlearning. 
We highlight results in \blue{blue} if the unlearning algorithm satisfies the criterion and highlight it in \orange{orange} otherwise.
%Results in \blue{blue} indicate criteria satisfaction, while results in \orange{orange} indicate failure to meet the criteria. 
For privacy leakage, large positive values suggest ~\orange{\scriptsize \textsf{over-unlearning}}, while large negative values suggest ~\orange{\scriptsize \textsf{under-unlearning}} (see \S\ref{subsec:metric}). This table covers the results for C1 to C4, while results for C5 and C6 are shown in \Cref{fig:deployer}.
%\nascomment{red/green are bad colors to use in contrast for colorblind readers.  blue/orange tends to work better I think.} \yang{updated} 
% \nascomment{can you add the annotations to the retrain row as well?}  \yang{added}
% \nascomment{explain missing values in caption.}
% %\nascommentif time, separate the colored annotations into their own columns, and right-justify numerical columns}\yang{added} %\name. 
% %\weijia{@yangsibo, explain the color}.
% \jaechan{I'll fill in the values that are currently missing}
 % \weijia{fix for $\TV$}
% \yang{Need to mention that this table is for C1 to C4, and results for C5 and C6 are in Fig xx.}
}
\label{tbl:main}
\centering
\small
\setlength{\tabcolsep}{8pt}
\renewcommand{\arraystretch}{0.75}
\resizebox{\linewidth}{!}{
\begin{tabular}{@{}lrlrlrlrl@{}}
\toprule
 & \multicolumn{2}{c}{\scriptsize \textbf{C1. No Verbatim Mem.}} & \multicolumn{2}{c}{\scriptsize \textbf{C2. No Knowledge Mem.}} & \multicolumn{2}{c}{\scriptsize \textbf{C3. No Privacy Leak.}}  & \multicolumn{2}{c}{\scriptsize \textbf{C4. Utiltiy Preserv.}} \\
 & \multicolumn{2}{c}{\scriptsize \textsf{VerbMem} on $\Du$ ($\downarrow$)} & \multicolumn{2}{c}{\scriptsize \textsf{KnowMem} on $\Du$ ($\downarrow$)} & \multicolumn{2}{c}{\scriptsize \textsf{PrivLeak} ($\in [-5\%, 5\%]$)} & \multicolumn{2}{c}{\scriptsize \textsf{KnowMem} on $\Dr$ ($\uparrow$)} \\
% \textbf{Method} & \textbf{Knowledge} & \textbf{Verbatim} & \multicolumn{2}{c}{\textbf{Privacy Leakage}}  & \textbf{Utility Preservation} \\
 % & $\downarrow$ & $\downarrow$ & Over-Unlearn $\downarrow$ & Under-Unlearn $\uparrow$ \\
\midrule
\multicolumn{9}{c}{\cellcolor[HTML]{EFEFEF}\textbf{\news}} \\ 
Target $\ft$&
$58.4$& &
$63.9$& &
$-99.8$& &
$55.2$ \\
Retrain $\fr$&
$\mathbf{20.8}$& &
$\mathbf{33.1}$& & 
$\mathbf{0.0}$& &
$\mathbf{55.0}$ \\
\midrule
\GA&      
$0.0$& $\da{100\%}$&
$0.0$& $\da{100\%}$&
$5.2$& ~\orange{\scriptsize \textsf{over-unlearn}}&
$0.0$& $\dar{100\%}$\\
\GD&
$4.9$& $\da{76.5\%}$&
$31.0$& $\da{6.3\%}$&
$108.1$& ~\orange{\scriptsize \textsf{over-unlearn}}&
$27.3$& $\dar{50.3\%}$\\
\GKL&
$27.4$& $\ua{31.4\%}$&
$50.2$& $\ua{51.5\%}$&
$-96.1$& ~\orange{\scriptsize \textsf{under-unlearn}}&
$44.8$& $\dar{18.5\%}$\\
\NPO&
$0.0$& $\da{100\%}$&
$0.0$& $\da{100\%}$&
$24.4$& ~\orange{\scriptsize \textsf{over-unlearn}}&
$0.0$& $\dar{100.0\%}$\\
\NPOD&
$1.2$& $\da{94.4\%}$&
$54.6$& $\ua{64.8\%}$&
$105.8$&~\orange{\scriptsize \textsf{over-unlearn}}&
$40.5$ & $\dar{26.3\%}$\\
\NPOKL&
$26.9$& $\ua{29.0\%}$&
$49.0$& $\ua{48.1\%}$& 
$-95.8$& ~\orange{\scriptsize \textsf{under-unlearn}}&
$45.4$& $\dar{17.4\%}$\\
\TV&
$57.2$& $\ua{174.7\%}$&
$66.2$& $\ua{100.0\%}$&
$-99.8$& ~\orange{\scriptsize \textsf{under-unlearn}}&
$55.8$& $\uar{1.5\%}$\\
$\WHP$&
$19.7$& $\da{5.6\%}$&
$21.2$& $\da{35.9\%}$&
$109.6$& ~\orange{\scriptsize \textsf{under-unlearn}}&
$28.3$& $\dar{48.5\%}$\\
\midrule
\multicolumn{9}{c}{\cellcolor[HTML]{EFEFEF}\textbf{\books}} \\
Target $\ft$&
$99.8$& &
$59.4$& &
$-57.5$& &
$66.9$\\
Retrain $\fr$&
$\mathbf{14.3}$& &
$\mathbf{28.9}$& &
$\mathbf{0.0}$& & 
$\mathbf{74.5}$\\
\midrule
\GA&      
$0.0$& $\da{100\%}$&    
$0.0$& $\da{100\%}$&    
$-25.0$& ~\orange{\scriptsize \textsf{under-unlearn}}&    
$0.0$& $\dar{100\%}$\\
\GD&      
$0.0$& $\da{100\%}$&    
$0.0$& $\da{100\%}$&  
$-26.5$&~\orange{\scriptsize \textsf{under-unlearn}}&   
$10.7$& $\dar{85.6\%}$\\
\GKL&    
$16.0$& $\ua{11.4\%}$&   
$21.9$& $\da{24.4\%}$&
$-40.2$&~\orange{\scriptsize \textsf{under-unlearn}}&
$37.2$& $\dar{50.0\%}$\\
\NPO&
$0.0$& $\da{100\%}$&
$0.0$& $\da{100\%}$&
$-24.3$&~\orange{\scriptsize \textsf{under-unlearn}}&
$0.0$& $\dar{100\%}$\\
\NPOD&
$0.0$& $\da{100\%}$&
$0.0$& $\da{100\%}$&
$-30.8$& ~\orange{\scriptsize \textsf{under-unlearn}}&
$22.8$ & $\dar{69.4\%}$\\
\NPOKL&
$17.0$& $\ua{18.2\%}$&
$25.0$ & $\da{13.4\%}$&
$-43.5$& ~\orange{\scriptsize \textsf{under-unlearn}}&
$44.6$ & $\dar{40.1\%}$\\
\TV&
$99.7$& $\ua{595.0\%}$&   
$52.4$& $\ua{81.2\%}$&  
$-57.5$&~\orange{\scriptsize \textsf{under-unlearn}}&   
$64.7$& $\dar{13.1\%}$\\
\WHP&
$18.0$& $\ua{25.2\%}$&
$55.7$& $\ua{92.9\%}$& 
$56.5$& ~\orange{\scriptsize \textsf{over-unlearn}}&
$63.6$& $\dar{14.6\%}$\\
\bottomrule \\ 
\end{tabular}}

\end{table}

\textbf{Retrained and target models.} 
We start with a general pretrained base model $f_0$, and finetune two models: $\ft$ on $\Du\cup\Dr$, and $\fr$ on $\Dr$ only. See \Cref{app:exp_setup} for details about finetuning. For each unlearning algorithm $\mathcal{U}$, we further generate the unlearned model $\fu=\mathcal{U}(\ft, \Du, \Dr)$. We ensure that $f_0$ has no access to $\Du,\Dr,\Dh$. Therefore, for \news, we use $f_0=\text{LLaMA-2 7B}$~\citep{llama2}, which was released \emph{before} the BBC news articles we use to construct our benchmarks; and for \books, we use $f_0=\text{ICLM-7B}$~\citep{shi2024incontext}, which does \emph{not} contain the Harry Potter books in its pretraining data.

\textbf{Unlearning experimental configuration.}
Following prior work \citep{Maini2024TOFUAT}, we run  \GA, \NPO, and their regularized variants using the AdamW optimizer~\citep{loshchilov2017decoupled} with a constant learning rate of $10^{-5}$ and a batch size of 32.
% To evaluate these methods' best-case performances, 
We employ the stopping criteria as follows:
if the utility (i.e., \textsf{KnowMem} on $\Dr$) of a model undergoing unlearning drops below that of $\fr$ within 10 epochs of unlearning, we stop at the first epoch where this condition holds; otherwise, we take a checkpoint from the 10th epoch.
For \TV and \WHP, to obtain the reinforced model for unlearning, we fine-tune the target model for 10 epochs using the same learning rate and batch size.
% Similar to the stopping criteria applied to \GA/\NPO variants, we evaluate the performance of \TV and \WHP by gradually increasing $\alpha$ from $2^0, 2^1, ..., 2^9$ and take the first $\alpha$ such that its utility drops below that of $\fr$.
Further details on the model fine-tuning and unlearning can be found in \Cref{app:exp_setup}.
% \yang{Is it possible to provide justification for these hyperparamaters?}

% \textbf{Metric Choices} \weijia{mention mia we use minkprob}

\subsection{Results: Data Owner Expectations}
\label{subsec:results-users}
 
We first analyze how eight unlearning methods meet data owner expectations (C1, C2 \& C3 in \S\ref{subsec:metric}).

% \Cref{tbl:main} summarizes 
% For comparison, the retained model sets an oracle performance that we wish the unlearned models to be similar to.
% Target models $\ft$ achieve high verbatim memorization rates: 93.3\% for the News dataset and 76.1\% for the Books dataset. Most unlearning methods, except for XX, 

% \textbf{C1+C2. Most methods are effective for unlearning memorization.}
% For both \news and \books, most of the unlearning methods drastically reduce verbatim and knowledge memorization in the target models, often surpassing the performance of the retrained models.
% Some methods, such as \GA and \NPO, achieve a score of 0 in both verbatim and knowledge memorization, completely preventing the model from producing any text related to the forget set.
% We later demonstrate that models with a substantial margin of forgetting often experience compromised utility on the retain set.

\textbf{C1\&C2. Most methods are effective for unlearning memorization.} As shown in \Cref{tbl:main}, most unlearning methods perform exceptionally well in [C1. No verbatim memorization] and [C2. No knowledge memorization], often reducing $\textsf{VerbMem}$ and $\textsf{KnowMem}$ even beyond the levels achieved by the retrained model. Notably, some methods, such as \GA and \NPO, achieve a score of 0 for both $\textsf{VerbMem}$ and $\textsf{KnowMem}$, meaning that these methods completely prevent the unlearned models from producing any text related to the forget set. However, as we will see later, these reductions often come at the cost of significant utility loss on the retain set.

% Most unlearning methods can significantly decrease the verbatim and knowledge memorization of the target models to a negligible level, sometimes even surpassing the performance of the retrained model. For instance, \NPOKL effectively reduces verbatim memorization by 79.9\% and knowledge memorization by XX\% on the BBC dataset. It is worth noting that some methods, such as $\GA$ and $\NPO$, cause the verbatim and knowledge memorization to drop to 0, likely due to the model collapsing and losing its utility.
 % \weijia{fix task vector}
% \weijia{what other finding to talk about here}
% \yang{`familiarity' and `unfamiliarity' seem a bit unnatural. Any better wording?}
% \yang{Also, are them any general trends we could report for these methods, say what type of methods/regularizer tend to result in under-unlearn, and what tend to result in over-unlearn?}

\begin{figure}[t]
% \vspace{-7mm}
\centering
\includegraphics[width=1\linewidth]{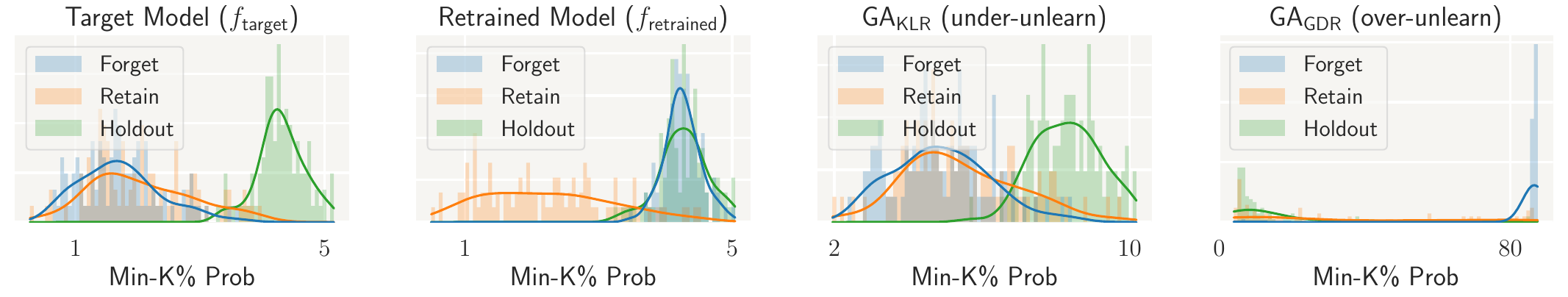}
\caption{\textbf{Distribution of Min-K\% Prob, an MIA metric, for $\Du$, $\Dh$, and $\Dr$.} Consistent with  the expected pattern in \Cref{fig:mia}, $\fr$ shows perfect unlearning, with the overlapping distributions for $\Du$ and $\Dh$. Existing approximate unlearning methods typically either under-unlearn or over-unlearn.  For example, \GKL shows slight under-unlearning, while \GD over-unlearns, pushing the Min-K\% Prob of $\Du$ to an extreme level.
%\weijia{(1) connect it to figure 2, use the same colors.. 
%(2) change the order of GA/WHP (3) change Loss to Min-K prob 
% (4) the overunlearn situation is bit tricky as it seems that the distribution between forget and holdout overlaps? maybe we can say it is model collapse instead of overunlearn? should we introduce the concept of model collapse somewhere in sec 3 (5) should we refer to Minkprob or just MIA metric in general in the experiment section}
 }
% \weijia{(1) connect it to figure 2, use the same colors.. (2) change the order of GA/WHP (3) change Loss to Min-K prob}\yang{Okay I will tweak colors later}
\label{fig:hist}
\end{figure}

\begin{figure}[t]
\vspace{-3mm}
\centering
\begin{minipage}{0.47\linewidth}
\centering
\includegraphics[width=\linewidth]{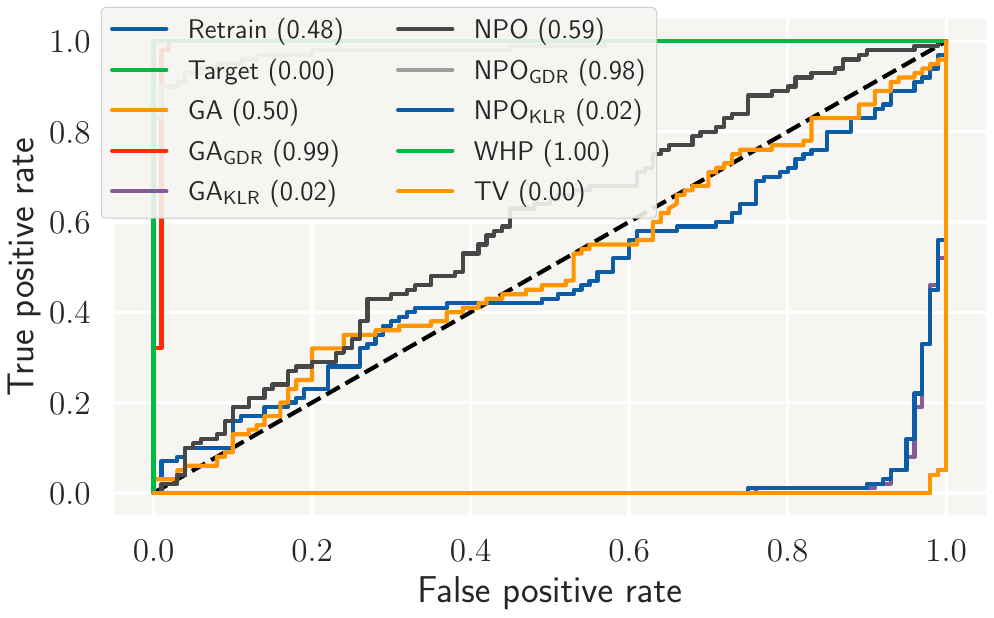}
\vspace{-5mm}
\caption{\textbf{ROC curves for $\Du$ vs. $\Dh$ on \news using Min-K\% Prob, with \textsf{AUC} scores in parentheses.} $\textsf{AUC}$$\approx$0.5 (i.e., $\fr$) means no significant distribution difference between two sets (i.e., no membership leakage). Most unlearning methods show under-unlearn ($\textsf{AUC}$$\ll$0.5) or over-unlearn ($\textsf{AUC}$ $\gg$0.5).
}
\label{fig:auc}
\end{minipage}
\hfill
\begin{minipage}{0.49\linewidth}
\centering
\includegraphics[width=\linewidth]{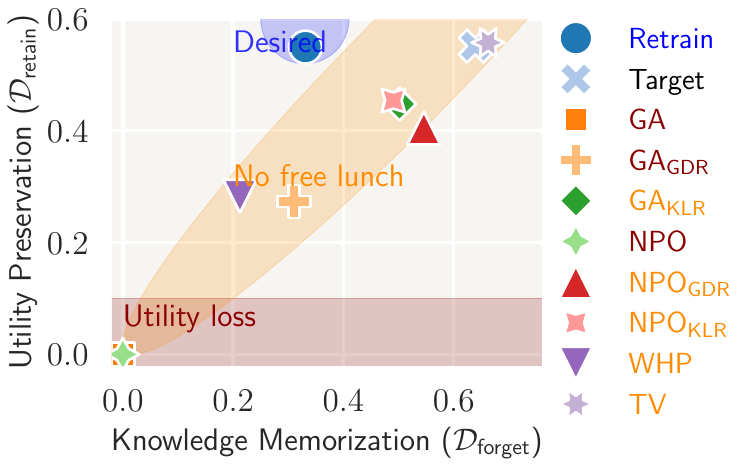}
\vspace{-5mm}
    \caption{
    \textbf{Utility preservation vs. knowledge memorization on BBC.} $\fr$ maintains high utility on $\Dr$ while showing low knowledge memorization on $\Du$. 
    \GA and \NPO without regularizers show significant utility loss, collapsing to the origin.
    Every other unlearning method unlearns the knowledge on $\Du$ at the cost of utility.}
    \label{fig:tradeoffs}
\end{minipage}
\end{figure}

\textbf{C3. Unlearning leads to privacy leakage.} Most unlearning methods reveal the membership of $\Du$ in $\Dt$ through under-unlearning ($\textsf{PrivLeak} \ll 0$) or over-unlearning ($\textsf{PrivLeak} \gg 0$), as shown in \Cref{tbl:main}. We further examine the effectiveness of membership inference by plotting ROC curves in \Cref{fig:auc}. The deviation from the diagonal line indicates the attacker's advantage over random guessing. We observe that the Min-K\% Prob based attack achieves $\textsf{AUC}\approx 0$ on $\ft$, confirming its effectiveness. Meanwhile, the ROC curve for $\fr$ closely follows the diagonal line ($\textsf{AUC}=0.47$), suggesting that perfect unlearning ensures MIA is no more effective than random guessing. Among the approximate unlearning methods, \GA and \NPOD without regularizers consistently over-unlearn ($\textsf{AUC} > 0.7$), whereas \KLR-regularized methods (\NPOKL and \GKL) tend to under-unlearn and barely improve privacy leakage over $\ft$.
%We hypothesize that the latter is potentially due to overly strict regularization, . 
\WHP also deviates from the diagonal significantly.

In \Cref{fig:hist}, we further visualize the distribution of Min-K\% Prob, the MIA metric computed across $\Du$, $\Dr$, and $\Dh$.
The behavior of $\ft$ and $\fr$ mirrors the patterns sketched in \Cref{fig:mia}, where $\Du$ and $\Dr$ are distinguishable in $\ft$ but overlap in $\fr$.  Existing approximate unlearning methods typically either under-unlearn or over-unlearn. For example,
\GKL does not sufficiently increase the Min-K\% Prob metric for $\Du$ to align with the distribution of $\Dh$, indicating under-unlearning. On the other hand,  \NPOD over-unlearns, significantly raising the MIA metric across all datasets and especially for $\Du$.
\subsection{Results: Deployment Considerations}
\label{subsec:results-deployer}

\textbf{C4. Unlearning significantly degrades model utility.}
\Cref{tbl:main} [C4 Utility Preserv.] shows that all unlearning methods compromise the model's utility by $24.2\%\sim 100\%$. Notably, several methods (\GA, \GD, \NPOD) lead to complete utility loss, rendering the unlearned models practically unusable. \Cref{fig:tradeoffs} illustrates the trade-offs between utility preservation on $\Dr$ and knowledge memorization on $\Du$. An ideal unlearned model should mimic the behavior of $\fr$ (desired region) by achieving a low level of memorization on $\Du$ while maintaining its utility. However, most methods, such as \GKL, \NPOKL, and \WHP, unlearn the knowledge on $\mathcal{D}_U$ at the cost of utility.
% showing that there is no free lunch in the unlearning process. \chiyuan{This last sentence sounds like suggesting good unlearning can never be achieved.}

% \textbf{C4. Unlearning degrades model utility significantly.}
% As shown in \Cref{tbl:main}, all the unlearning methods compromise the model’s utility  [C4 Utility Preserv.] by 24.2\% to 100\%. Notably, Several unlearning methods, such as \GA, \GD and \NPOD, leads to complete utilty loss, showing the unlearned models practically unusable. 

% \Cref{fig:tradeoffs} shows the trade-offs between utlity preservation on $\Dr$ and knowledge memorization on $\Du$.
% An ideal unlearned model should mimic the behavior of $\fr$ (desired region) by achieving a low level of memorization on $\Du$ while maintaining its utility on $\Dr$.
% However, most methods such as \GKL, \NPOKL and \WHP, unlearn the knowledge on $\Du$ at the cost of utility (No free lunch).

% Some methods such as 
% Notably, the unregularized methods \GA and \NPO reduce the model’s utility to zero in both the books and news settings, rendering the model practically unusable.
% We posit that these methods exploit their losses by indiscriminately forgetting all the knowledge they contain.

\begin{wrapfigure}{r}{0.5\textwidth}
    \centering
    % \vspace{-4mm}
    \begin{subfigure}[t]{\linewidth}
        \centering
        \includegraphics[width=\linewidth]{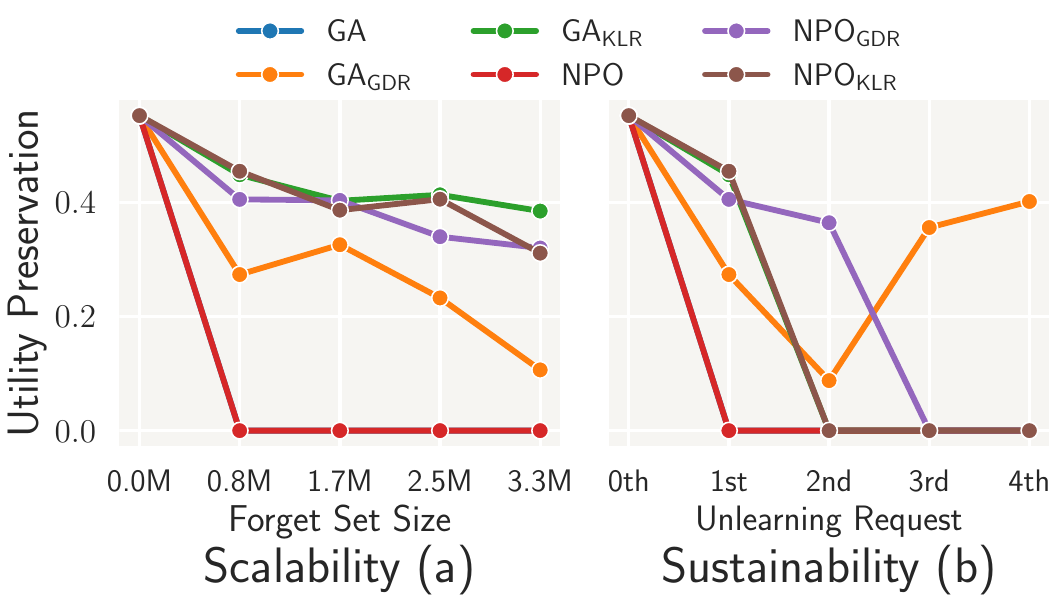}
        % \vspace{-6mm} 
    \end{subfigure}
    % \begin{subfigure}[t]{0.48\linewidth}
    %     \centering
    %     \includegraphics[width=\linewidth]{figures/c6.pdf}
    %     \vspace{-6mm}
    %     \caption{Sustainability}
    % \end{subfigure} 
    % \vspace{-2mm}
    \caption{\textbf{The performance of \GA, \NPO, and their regularized variants, measured by utility preservation, degrades with larger forget set sizes (a) and sequential unlearning requests (b).}
    % In all points shown in the plot, the unlearning algorithms reduced verbatim memorization to below 10, outperforming $\fr$.
    % \weijia{for scalability plot, show the number for berfore unlearning? (forget size=0)}
    % \yang{I feel the ``best unlearning methods'' is quite vague here... any thoughts on how to be more concrete?} \weijia{agree, why we don't choose the GA-KLR and NPO-KLR methods? as GA-GDR nad NPO-GDR make the utility drop to 0?}  
    % \jaechan{But those methods are under-unlearned even on the largest dataset so I don't think their trends on smaller datasets are that interesting... also their margin of downfall on utility isn't huge.
    % GA-GDR and NPO-GDR have a huge downfall in utility while still not having collapsed according to C3}
    %(a) shows the trend of utility preservation with respect to different forget set sizes. (b) shows the trend of utility preservation with respect to sequential unlearning requests, where the unlearning process is applied multiple times to the model. 
    }
    \label{fig:deployer}
    \vspace{-4mm}
\end{wrapfigure}
\textbf{C5. Unlearning methods scale poorly with forget set sizes.}
% Compared with prior work \citep{Maini2024TOFUAT}, our forget set size in \news is $30\times$ larger (3.3 million tokens). We observed that in our setting, most methods' utility preservation is poor. Therefore, we perform a scalability analysis to identify at which point or scale they start to fail.
To evaluate the robustness of the unlearning methods to larger forget sets, we collect additional news articles from the same distribution  to scale our \news corpus from 0.8M tokens to 3.3M tokens and observe the utility preservation at four different forget set sizes.
As shown in \Cref{fig:deployer} (a), the model utility decrease with the size of the forget set and achieves a minimum at the largest size. 

\textbf{C6. Unlearning methods cannot sustainably accommodate sequential unlearning requests. }
% \weijia{@jaechan, why they are the best?, we need better justification}
To evaluate the robustness of these unlearning methods to more than one unlearning requests, we sequentially apply $k$ unlearning processes, each with respect to a different forget set.
To simulate sequential unlearning, we partition the extended \news forget set (comprised of 3.3M tokens) into four disjoint folds (each containing 0.8M tokens) and apply the unlearning methods to each fold in a sequential manner.

We again select utility preservation as the target metric for comparison.
As shown in Figure \ref{fig:deployer} (b), the performance of an unlearned model tends to decrease significantly with respect to the number of unlearning requests, indicating that current unlearning methods are not yet ready to handle sequential unlearning in a sustainable manner.

% \textbf{Unlearning methods are more scalable than they are sustainable.}
% The two plots in Figure \ref{fig:deployer} are directly comparable to each other, as the cumulative data expected to be forgotten at each point in the $x$-axis is the same for both scalability and sustainability setups.
% For example, a model after unlearning the 0.8M-token-large $\Du$ and the first two unlearning requests are expected to forget equivalent data.
% We observe that all the unlearning methods in the sustainability setup with the exception of \GD collapse to zero utility, whereas most of their counterparts in the scalability setup yield nonzero utility.
% In general, unlearning $\Du$ that is split into multiple sequential requests leads to a more drastic decrease in utility preservation than unlearning $\Du$ once.

% \input{content/5-vision}
% \newpage
\section{Related Work}
\label{app:related}

\paragraph{Machine unlearning for non-language model applications.}
Machine unlearning is a long-running, well-studied topic.
Several studies have explored exact unlearning, aiming to make the unlearned model ($\fu$) exactly identical to the reference model ($\fr$). 
As expected, this can only be accomplished in simple models like SVMs~\citep{cauwenberghs2000incremental,tveit2003incremental,romero2007incremental,karasuyama2010multiple} or naive Bayes models~\citep{cao2015towards}.
Another approach is to ensure that the unlearned model $\fu$ is probabilistically indistinguishable from $\fr$~\citep{ginart2019making,guo2020certified}, and this view of certifiable unlearning is closely related to differential privacy~\citep{dwork2006calibrating, dwork2006our}.
This rigorous definition of unlearning has inspired several theoretical works that characterize the feasibility of unlearning in convex and non-convex models, but those proposed algorithms are too computationally costly to operate on modern-day LMs~\citep{izzo2021approximate,neel2021descent,ullah2021machine,sekhari2021remember,gupta2021adaptive}.
%Notable examples include \cite{cauwenberghs2000incremental, tveit2003incremental, romero2007incremental, karasuyama2010multiple}. These works are often supported by strong theoretical foundations. However, due to their restrictive requirements, exact unlearning is typically feasible only in simpler models like SVM.
% There is a line of works studying how to do exact unlearning (make $\fu$ exact the same as $\fr$ in our notation), for example, \cite{cauwenberghs2000incremental,tveit2003incremental,romero2007incremental,karasuyama2010multiple}.
% These works usually have theoretical support.
% However, due to the restrictive requirements, exact unlearning may only be achieved in some simple models, like SVM.
Several more tractable unlearning algorithms have been proposed \citep{borkan2019nuanced, ginart2019making, thudi2022unrolling, chourasia2023forget} with broader applications such as image classification \citep{ginart2019making,golatkar2020eternal}, text-to-image generation \citep{gandikota2023erasing,zhang2023forget,fan2023salun}, Federated Learning \citep{liu2020federated,che2023fast,halimi2022federated, huang2022dataset} and Recommender Systems \citep{li2024making}.

% \yang{TODO: add the CUT paper from Scale as other applications of unlearning}

\paragraph{Machine unlearning for language models: methods and applications.}
Machine unlearning has recently found its way into language model applications. In \S\ref{sec:method}, we discuss some standard unlearning methods based on parameter optimization, like the Gradient Ascent and its variance. Other notable non-training-based unlearning methods include localization-informed unlearning \citep{meng2022locating,wu2023depn,wei2024assessing}, which involves identifying model units (e.g., layers, neurons) closely related to the unlearning data or tasks and then locally editing and modifying the units. In-context unlearning \citep{pawelczyk2023context} offers another approach, treating the model as a black box and modifying its output results using external knowledge. 

Machine unlearning has also been applied to various downstream language model tasks, though the unit of machine unlearning may differ from what we study in this work. Our evaluation focuses on unlearning specific examples or datasets, aiming to make LMs forget either the phrasing or the content knowledge of targeted data, while preserving their utility for data not targeted for removal. This is crucial for ensuring privacy and copyright compliance. In addition to this specific unlearning, there’s also a broader application similar to model editing, where outdated information is replaced with new knowledge~\citep{pawelczyk2023context, yu2023unlearning, belrose2024leace}. Moreover, efforts have been made to eliminate harmful behaviors in language models by creating toxicity benchmarks and enhancing safety measures~\citep{lu2022quark, yao2023large, li2024wmdp, zhang2024negative}.
Despite these varied approaches to unlearning at different operational and knowledge levels, the evaluation principles we propose such as preserving utility, ensuring scalability, and maintaining sustainability—are relevant across these contexts.

\paragraph{Machine unlearning for language models: evaluation.} Evaluating machine unlearning methods for language model applications is also critical. Most previous studies have focused this evaluation on specific tasks such as question answering or sentence completion. For example, \citet{eldan2023s} experiment with unlearning to forget Harry Potter books and demonstrate the effectiveness of their methods by showing that familiarity scores, measured through completion-based, token-probability-based, and question-answering evaluations, significantly decline post-unlearning. \citet{lynch2024eight} further suggest comparing unlearned models with perfectly retrained models. Their evaluation finds that while familiarity scores with the forget set may drop post-unlearning, they still remain higher than those of the retrained model. \citet{wei2024evaluating} evaluate the feasibility of using unlearning techniques to prevent language models from generating copyrighted content. The closest work to ours is TOFU~\citep{Maini2024TOFUAT}, a benchmark featuring 200 synthetic author profiles, each with 20 question-answer pairs, divided into forget and retain sets. However, TOFU is relatively small-scale (0.15M tokens) and focuses on the evaluation of question answering.
% lacks separation of original text and knowledge, which conflates the evaluation of unlearning verbatim text and knowledge (C1 and C2 in \name). 
Additionally, current evaluations focus on limited aspects of data owner expectations and do not adequately reflect real-world deployment considerations, such as scalability and potential sequential unlearning requests. In contrast, \name formally defines different unlearning scopes and corresponding metrics, resulting in a systematic six-way evaluation featuring both data owners' and deployers' expectations. The evaluation uses a large-scale corpus of over 6 million tokens, separated into verbatim text and knowledge sets. We also note that some of our findings align with previous evaluations. For example, our observation that over- or under-unlearn can exacerbate privacy leakage (\S\ref{subsec:results-users}) is consistent with the recent work by \citet{hayes2024inexact}. Our findings align with the the concurrent study by \citet{shumailov2024ununlearning}  showing that unlearning gives a false sense of security as unlearned knowledge can resurface through in-context learning.

\paragraph{Survey papers.} We direct readers to several insightful survey papers for further reading. For non-LLM applications, notable surveys include \citet{shintre2019making, nguyen2022survey, thudi2022unrolling, xu2023survey}. Additionally, the NeurIPS 2023 machine unlearning competition for image classification\footnote{\url{https://unlearning-challenge.github.io}} is a valuable source of empirical methods tailored for this specific application \citep{neurips-2023-machine-unlearning}. For language model applications, \citet{si2023knowledge} categorize unlearning methods into different families and summarize datasets for evaluating unlearning. \citet{liu2024rethinking} review LM unlearning algorithms by targets and methods, discuss the effectiveness and efficiency of existing approaches and emphasize the importance of clearly defining the unlearning scope.
\vspace{-2mm}
\section{Conclusion}
\label{sec:conclusion}
% \vspace{-4mm}

In this work, we propose \name, a comprehensive machine unlearning evaluation benchmark that highlights six desirable properties from the perspectives of both data owners and model deployers. We find that current unlearning methods successfully prevent the model's memorization of content at a significant cost to utility on data not intended for removal. They also lead to severe privacy leakage and cannot sustainably accommodate successive unlearning requests or large-scale content removal. These findings highlight the need for future research into more robust unlearning methods.
% to develop better machine unlearning methods.

\textbf{Limitations.} 
While \name provides a systematic benchmark for evaluating unlearning algorithms, it does not consider all possible considerations.
% based on criteria desired by data owners and model deployers. However, we are not able to encompass all possible considerations. 
For example, data owners may have additional expectations, such as ensuring their information cannot be probed from intermediate activations~\citep{song2020information} or receiving formal guarantees of unlearning success~\citep{sekhari2021remember, gupta2021adaptive, ghazi2023ticketed}. Similarly, deployers may expect other capabilities, like fine-tuning and in-context learning, to be preserved, and may prefer unlearning algorithms that are both computationally efficient and storage-wise cheap (e.g. does not need to keep a copy of the retain set).
\name currently evaluates unlearning for language models using books and news articles, but it could be extended to other corpora, such as medical notes~\citep{johnson2016mimic, johnson2020mimic} and emails~\citep{klimt2004enron}, which often involve privacy concerns~\citep{li2023multi, huang2023privacy}. We also plan to evaluate different-sized LMs in the future. Finally, our approach can be generalized to construct multi-faceted benchmarks for  multimodal models~\citep{golatkar2020forgetting, cheng2023multimodal, zhang2024unlearncanvas}. Further
discussion on broader impact are in Appendix \ref{app:broader_impact}.

\section{Acknowledgements}
We thank Eric Wallace, Robin Jia, Howard Chen, and anonymous reviewers of the GenLaw workshop for the valuable feedback and discussions.

\clearpage
\bibliographystyle{neurips_2024}
\bibliography{misc/main}

% \clearpage
% \newpage
% \input{misc/checklist_data}

\newpage
\appendix
\appendixpage
\startcontents[sections]
\printcontents[sections]{l}{1}{\setcounter{tocdepth}{2}}
\newpage

\section{Broader Impact}
\label{app:broader_impact}

As LMs are deployed broadly and publicly, there is mounting legal and social pressure on deployers to release models that permit effective unlearning when requested by data owners~\citep{GDPR2016a,githublitigation,openailawsuit2}.
These incentives have prompted a flurry of new unlearning algorithms stemming from different technical perspectives.
As such, systematic evaluation of the strengths and weaknesses of these methods when executing realistic unlearning requests on popular models is essential. 
\name disentangles several desirable properties of unlearning algorithms and finds that no existing algorithm is able to satisfy all of the data owner and deployer considerations.
We hope that our fine-grained, multi-faceted framework facilitates the improvement of unlearning algorithms.
Moreover, we expect that the general approach of designing metrics to balance the considerations of various stakeholders is flexible and can adapt to the rapidly shifting legal, social, and economic landscape.

% \weijia{talk about negative impacts }
%\yang{impact on fairness? unrepresentative groups}

We also acknowledge the potential negative impacts of our study. One limitation of our evaluation benchmark is that we do not have comprehensive study of how unlearning would impact the model performance for different user bases, especially underrepresented groups. However, we note proper handling and evaluation of fairness issues in unlearning is still an active ongoing research area~\citep{zhang2024forgotten,oesterling2024fair}, therefore we leave it as future work. Additionally, our work may be misinterpreted towards skepticism regarding the broader use of machine unlearning, as our current evaluation reveals that existing unlearning methods are not yet ready for effective real-world deployment. However, machine unlearning, especially for large language models, is a young and active research area and new algorithms are constantly being proposed. We emphasize that our results is not a criticism of the paradigm of machine unlearning, but a study of the potential downsides of existing methods and a call for better algorithms. We believe our benchmark is an important step towards guiding future algorithm design of machine unlearning research towards more realistic deployment scenarios.

\newpage
\section{Experimental Details}
\label{app:exp_details}

\subsection{Compute Configurations}
\label{app:exp_compute}

All experiments are conducted on 8 NVIDIA A40 GPU cards in a single node.
% We conduct all the experiments on \yang{Hardware specification...}

\subsection{Experimental Setup}
\label{app:exp_setup}

\paragraph{Finetuning details.}
As described in \S \ref{subsec:exp_setup}, for \news, we start from $f_0=\text{LLaMA-2 7B}$~\citep{llama2} and finetune the model on the BBC news articles for 5 epochs with a constant learning rate of $10^{-5}$ and a batch size of 32 . For \books, we start from $f_0=\text{ICLM 7B}$~\citep{llama2} and finetune the model on the Harry Potter books with same set of hyperparameters.

\paragraph{Unlearning details.}
For all the unlearning methods in Table \ref{tbl:main}, we use a constant learning rate of $10^{-5}$ and a batch size of 32.
For $f_\text{reinforced}$ used in \WHP and \TV, we fine-tune $\ft$ for 10 epochs.

Before evaluation, for each unlearning method, we select its optimal epoch or $\alpha$ (both of which are parameters that control a degree of unlearning) by using our unlearning stopping criteria based on the unlearned model's utility on $\Dr$ compared to that of $\fr$.
The chosen epochs or $\alpha$'s for each method are listed below.
\begin{table}[ht]
\small
\caption{Optimal epochs or $\alpha$'s for each unlearning method.}
\label{tab:epoch}
\centering
\begin{tabular}{c|c| c}
\toprule
\textbf{Unlearning Method} & \news & \books
\\
\midrule
\GA & epoch 1 & epoch 1 \\
\GD & epoch 7 & epoch 1\\
\GKL & epoch 10 & epoch 5 \\
\NPO & epoch 1 & epoch 1 \\
\NPOD & epoch10 & epoch 1 \\
\NPOKL & epoch 10 & epoch 4 \\
\TV & $\alpha = 2^9$ & $\alpha = 2^9$ \\
\WHP & $\alpha = 2^2$ & $\alpha = 2^8$ \\
\bottomrule
\end{tabular}
\end{table}

\subsection{Efficiency of Unlearning Methods}
\label{app:efficiency}
We report the efficiency of unlearning methods in \Cref{tab:time}, measured by the wall-clock time for a single gradient update step of unlearning. The time measurements were conducted using 8 NVIDIA A40 GPUs on a single node, with a batch size of 32 and an input length of 2048 tokens.
Each step corresponds to one gradient update processing a total of 65,536 tokens (32 $\times$ 2048 tokens). For \TV and \WHP, each step represents one iteration of fine-tuning to create the reinforced model.

% For \TV, we do not discuss the time required to construct a task vector from $\ft$ and $f_\text{reinforced}$, as it is negligible compared to the time needed to create $f_\text{reinforced}$. Additionally, we note that \WHP, which achieves unlearning by interpolating the output logits of $\ft$ and $f_\text{reinforced}$, requires roughly double the inference-time computation and disk storage compared to models unlearned using other methods.
% We report the efficiency of unlearning methods in \S\ref{sec:method} in wall-clock time taken for running one step (one gradient update) of unlearning/fine-tuning.
% The time is measured with 8 NVIDIA A40 GPUs in a single node with a batch size of 32 and the input length of 2048 tokens.
% Each step signifies one gradient update of direct unlearning for \GA, \NPO, and their regularized variants ($32x2048=65K$ tokens); and one iteration of fine-tuning (in creating the reinforced model) for \TV and \WHP.

% For \TV, we omit the discussion on the time it takes to construct a task vector out of $\ft$ and $f_\text{reinforced}$, which is negligible compared to the time taken for creating $f_\text{reinforced}$ itself.
% We also note that \WHP, which achieves unlearning by interpolating the output logits of $\ft$ and $f_\text{reinforced}$, requires roughly twice the amount of inference-time computation and disk storage compared to models unlearned using other methods.
% \weijia{2048token*32: }
\begin{table}[ht]
\small
\caption{Wall-clock time required for each unlearning method, measured in seconds per step.}
\label{tab:time}
\centering
\begin{tabular}{c|c}
\toprule
\textbf{Unlearning Method} & \textbf{Time (Seconds/Step)} \\
\midrule
\GA & 4.14 \\
\GD & 6.05 \\
\GKL & 7.58 \\
\NPO & 5.68 \\
\NPOD & 7.59 \\
\NPOKL & 9.11 \\
\TV & 4.14 \\
\WHP & 4.14 \\
\bottomrule
\end{tabular}
\end{table}
% \weijia{@jaechan add efficiency}
% \weijia{retain set 2}

\newpage
\section{More Experimental Results}
\label{app:results}

\subsection{Confidence Intervals for C1, C2 and C4 in \autoref{tbl:main}}
% We compute confidence intervals for C1, C2 and C4 (Mean ROUGE-L F1) (We didn't compute the confidence interval for C3 because .... \weijia{do we need to explaian this}). 
% We use bootstrapping\footnote{\url{https://docs.scipy.org/doc/scipy/reference/generated/scipy.stats.bootstrap.html}} to compute a two-tailed 95\% confidence interval for each of the mean Rouge-L score reported in \autoref{tbl:main}.
% For each interval to be computed, we draw 9,999 bootstrap resamples and use the ``percentage'' method
% (i.e., identify the values at the 2.5\% and 97.5\% percentiles from the sorted bootstrap distribution, which correspond to the lower and upper bounds of the interval, respectively).
We compute confidence intervals for C1, C2, and C4 (Mean ROUGE-L F1) using bootstrapping\footnote{\url{https://docs.scipy.org/doc/scipy/reference/generated/scipy.stats.bootstrap.html}}. 
%\weijia{}
%The confidence interval for C3 was not computed due to \weijia{[reason for not computing C3 confidence interval]}\yang{Actually I realized that we should still be able to report CI for AUC. But this may require a bit more complicated implementation (may not work with the off-the-shelf function). Let's discuss this in our meeting}. 
For each mean ROUGE-L score reported in \autoref{tbl:main}, we draw 9,999 bootstrap resamples and calculate a two-tailed 95\% confidence interval using the ``percentage'' method. 
% This method involves identifying the values at the 2.5% and 97.5% percentiles from the sorted bootstrap distribution, which correspond to the lower and upper bounds of the interval, respectively.
\begin{table}[th]
\small
\caption{
95\% confidence intervals computed for mean Rouge-L scores used in C1, C2, and C4. %\weijia{remove percentage}
}
\label{tbl:ci}
\centering
\small
\setlength{\tabcolsep}{12pt}
\resizebox{\linewidth}{!}{
\begin{tabular}{@{}lrlrlrl@{}}
\toprule
 & \multicolumn{2}{c}{\scriptsize \textbf{C1. No Verbatim Mem.}} & \multicolumn{2}{c}{\scriptsize \textbf{C2. No Knowledge Mem.}} & \multicolumn{2}{c}{\scriptsize \textbf{C4. Utiltiy Preserv.}} \\
 & \multicolumn{2}{c}{\scriptsize \textsf{VerbMem} on $\Du$ ($\downarrow$)} & \multicolumn{2}{c}{\scriptsize \textsf{KnowMem} on $\Du$ ($\downarrow$)} & \multicolumn{2}{c}{\scriptsize \textsf{KnowMem} on $\Dr$ ($\uparrow$)} \\
\midrule
\multicolumn{7}{c}{\cellcolor[HTML]{EFEFEF}\textbf{\news}} \\ 
Target $\ft$&
$58.4$& [54.1, 62.9]&
$63.9$& [58.7, 69.0]&
$55.2$& [50.7, 59.9]\\
Retrain $\fr$&
$\mathbf{20.8}$& [18.5, 23.7]&
$\mathbf{33.1}$& [26.8, 39.5]& 
$\mathbf{55.0}$& [50.3, 59.8]\\
\midrule
\GA&      
$0.0$& [0.0, 0.0]&
$0.0$& [0.0, 0.0]&
$0.0$& [0.0, 0.0]\\
\GD&
$4.9$& [4.5, 5.2]&
$31.0$& [24.2, 38.0]&
$27.3$& [21.9, 33.0]\\
\GKL&
$27.4$& [25.1, 29.9]&
$50.2$& [43.1, 56.9]&
$44.8$& [39.2, 50.5]\\
\NPO&
$0.0$& [0.0, 0.0]&
$0.0$& [0.0, 0.0]&
$0.0$& [0.0, 0.0]\\
\NPOD&
$1.2$& [0.3, 2.3]&
$54.6$& [47.5, 61.5]&
$40.5$ & [34.7, 46.2]\\
\NPOKL&
$26.9$& [24.7, 29.3]&
$49.0$& [41.8, 61.5]& 
$45.4$& [39.8, 51.1]\\
\TV&
$57.2$& [52.6, 62.0]&
$66.2$& [61.3, 71.2]&
$55.8$& [51.0, 60.6]\\
\WHP&
$19.7$& [17.8, 21.6]&
$21.2$& [16.0, 26.7]&
$28.3$& [23.3, 33.4]\\
\midrule
\multicolumn{7}{c}{\cellcolor[HTML]{EFEFEF}\textbf{\books}} \\
Target $\ft$&
$99.8$& [99.8, 99.9]&
$59.4$& [52.7, 66.0]&
$66.9$& [59.6, 73.8]\\
Retrain $\fr$&
$\mathbf{14.3}$& [13.6, 15.1]&
$\mathbf{28.9}$& [22.1, 35.7]&
$\mathbf{74.5}$& [68.4, 80.0]\\
\midrule
\GA&      
$0.0$& [0.0, 0.0]&    
$0.0$& [0.0, 0.0]&     
$0.0$& [0.0, 0.0]\\
\GD&      
$0.0$& [0.0, 0.0]&    
$0.0$& [0.0, 0.0]&    
$10.7$& [6.2, 15.7]\\
\GKL&    
$16.0$& [14.8, 17.2]&   
$21.9$& [16.4, 27.7]&
$37.2$& [29.5, 45.0]\\
\NPO&
$0.0$& [0.0, 0.0]&
$0.0$& [0.0, 0.0]&
$0.0$& [0.0, 0.0]\\
\NPOD&
$0.0$& [0.0, 0.0]&
$0.0$& [0.0, 0.0]&
$22.8$& [16.1, 30.1]\\
\NPOKL&
$17.0$& [15.7, 18.2]&
$25.0$& [19.0, 31.5]&
$44.6$& [36.5, 52.8]\\
\TV&
$99.7$& [99.6, 99.8]&   
$52.4$& [45.0, 59.7]&  
$64.7$& [57.1, 71.8]\\
\WHP&
$18.0$& [16.4, 19.7]&
$55.7$& [48.6, 62.8]& 
$63.6$& [56.3, 70.9]\\
\bottomrule \\ 
\end{tabular}}

\end{table}

% The computed confidence intervals for C1, C2, and C4 all fall within $\pm 10$\% of their respective mean ROUGE-L scores. This observation suggests that the per-sample ROUGE-L scores provide a relatively precise representation of the mean scores presented in \autoref{tbl:main} and discussed earlier. \weijia{not sure what is the takeaway here?}
% The narrow range of the confidence intervals indicates that the mean scores are reliable estimates of the models' performance, and the individual sample scores do not deviate significantly from these means. This finding reinforces the validity of the conclusions drawn from the mean ROUGE-L scores and supports the robustness of the evaluation methodology employed in this study.
% We observe that all the intervals computed are within $\pm 10$\% from the mean, indicating that each per-sample Rouge-L score provides a relatively precise representation of the mean scores presented in \autoref{tbl:main} and above.

% \weijia{add a paragraph discussing the experimental setting and how to interpret results}

% \input{app/dataset_info}
\newpage
\section{Dataset Details}
\label{app:data_details}

\textbf{GPT-generated QA pairs.} 
% \yang{Please provide more details for how the QA are collected (e.g. the GPT4 endpoint, prompting template etc).}
% As discussed in \S\ref{subsec:dataset} and shown in \Cref{tbl:dataset}, for each corpus in \name, we use GPT-4 to automatically generate \yellow{\small Knowledge} set: a set of (question, answer) pairs for evaluating a model's knowledge memorization. 
% For both \news and \books, we use GPT-4 to automatically generate \yellow{\small Knowledge} set: a set of (question, answer) pairs for evaluating a model's knowledge memorization.
We begin the generation by partitioning the \white{\small Verbatim} text of each corpus into a set of 2048-token excerpts using LLaMA-2's tokenizer.
For each QA pair to generate, we randomly sample an excerpt from this set and prompt GPT-4 (\texttt{gpt-4o-2024-05-13}) to create a JSON object with two fields: ``\textit{question}'' (a question that can only be answered using specific information from the excerpt) and ``\textit{answer}'' (an answer to the ``question'' extracted verbatim from the excerpt).
We validate and exclude any pairs whose answers cannot be found verbatim in their corresponding excerpts.
This verbatim requirement ensures that our \yellow{\small Knowledge} set is used precisely to evaluate the model's ability to correctly associate questions with relevant portions of the training data.
% This ability is crucial especially in the application of unlearning for copyright-related applications,  where generating a verbatim extraction is more concerning than producing its paraphrase.
% \yang{I feel the last sentence might invite controversy, as it remains unclear whether paraphrased text would infringe on copyrighted materials. Maybe let's just skip it?}

For each QA pair to generate, we initiate a new conversation with GPT-4 with its corresponding excerpt. The instruction begins with a system prompt that specifies the desired format of generated QA pairs as follows:

\begin{example}[System Prompt for Generating QAs with GPT-4]
\small
You will be provided with an excerpt of text. Your goal is to create a question-answer pair that assesses reading comprehension and memorization, ensuring that the question can only be answered using details from the excerpt.

\quad

Please submit your response in a JSON format with the following fields:

- ``question'': A single question related to the excerpt. The question should be specific enough that it does not allow for an answer other than the one you provide. In particular, it should not be answerable based on common knowledge alone. Also, a few words extracted from the excerpt must suffice in answering this question.

- ``answer'': A precise answer extracted verbatim, character-by-character from the excerpt. The answer to this question must be short, phrase-level at most. The length of the extraction should be minimal, providing the smallest span of the excerpt that completely and efficiently answers the question.
\end{example}

We then present the excerpt as a user prompt to the model and collect the generated QA pairs. Here are two example generated QA pairs from the \yellow{\small Knowledge} set of \news:

\begin{example}[QA Pair Generated by GPT-4: Example \#1]
\small
\textbf{Excerpt (User prompt):} ...According to the Stockholm International Peace Research Institute (SIPRI), the US accounted for 69\% of Israel's arms imports between 2019 and 2023...

\textbf{Question:}
According to the Stockholm International Peace Research Institute (SIPRI), what percentage of Israel's arms imports between 2019 and 2023 came from the US?

\textbf{Answer:}
69\%
\end{example}

\begin{example}[QA Pair Generated by GPT-4: Example \#2]
\small
\textbf{Excerpt (User prompt):}
...Wednesday's event will be moderated by tech entrepreneur David Sacks, a close ally of the Tesla founder and a supporter of Mr DeSantis...

\textbf{Question:}
Who will moderate Wednesday's Twitter Spaces event featuring Mr DeSantis?

\textbf{Answer:}
tech entrepreneur David Sacks
\end{example}

% The generated QA pairs do not contain personal identifiable information or offensive content.
% \yang{@Jaechan, How to  justify this statement? Did you use some rule-based checkers?}

% \newpage
\textbf{Dataset segmentation.} 
% \weijia{@jaechan, explain the table 4}
% \yang{Let's add a table for the forget/retain/heldout split for both BBC and HP.}
% \Cref{tbl:corpus_comparison_app}  presents detailed statistics for \name. For both the \news and \books datasets, we include the type of documents along with the number of tokens in each dataset. Additionally, \name incorporates $\Dr^{(\text{reg})}$, a distinct retain set which is seen by $\ft$ but not included in $\Du$. This set is used exclusively with the \GDR and \KLR regularizers discussed in \S \ref{sec:method}. To ensure that regularized methods do not directly optimize towards the evaluation set $\Dr$ used in \S \ref{sec:exp}, $\Dr^{(\text{reg})}$ is kept disjoint from $\Dr$.
\Cref{tbl:dataset} shows examples from \name  and 
\Cref{tbl:corpus_comparison_app}  presents detailed statistics for \name. For both the \news and \books datasets, we include the type of documents along with the number of tokens in each dataset. Additionally, \name incorporates $\Dr^{(\text{reg})}$, a distinct retain set which is seen by $\ft$ but not included in $\Du$. This set is used exclusively with the \GDR and \KLR regularizers discussed. To ensure that regularized methods do not directly optimize towards the evaluation set $\Dr$ , $\Dr^{(\text{reg})}$ is kept disjoint from $\Dr$.

\begin{table}[ht]
\centering
% \small
\vspace{-3mm}
\setlength{\tabcolsep}{3pt}
\renewcommand{\arraystretch}{1.15}
\caption{\textbf{Statistics of the \name dataset.} Corpus sizes are reported in tokens, shown in (). Retain Set$_{\text{reg.}}$ is disjoint from the standard Retain Set used in evaluation and is employed in unlearning training to preserve utility through regularizers.
% \weijia{@jaechan can you add the stats here? }
}

\resizebox{\linewidth}{!}{
\begin{tabular}{m{1cm}m{4.5cm}m{4.5cm}m{4.5cm}m{4.5cm}}
\toprule
\textbf{Corpus} & \textbf{Forget Set} & \textbf{Retain Set} & \textbf{Retain Set$_{\text{reg.}}$} & \textbf{Holdout Set} \\
\hline
% \rowcolor{gray}
\news &
News Articles \color{gray}{(3.3M)} &
News Articles \color{gray}{(1.6M)} &
News Articles \color{gray}{(1.6M)} &
News Articles \color{gray}{(2.0M)} \\
\books &
Harry Potter Books \color{gray}{(1.1M)} &
Harry Potter FanWiki \color{gray}{(0.5M)} &
Harry Potter FanWiki \color{gray}{(0.2M)} &
Harry Potter Books \color{gray}{(0.6M)} \\
\bottomrule
\end{tabular}
}
\label{tbl:corpus_comparison_app}
\end{table}

% $\Du$ is the largest, followed by $\Dh$, with $\Dr$ being the smallest.
% We also note the existence of $\Dr^{(\text{reg})}$, a retain set (i.e., whose samples are seen by $\ft$ and not included in $\Du$) exclusively employed for \GDR and \KLR regularizers introduced in \S\ref{sec:method}.

% \textbf{Dataset License} We use BBC news and Harry Potter Books as our raw datasets. BBC news is licensed under MIT license, and Harry Potter Books is licensed under CC0 1.0 Universal.

% \yang{Data license }

\end{document}